\documentclass{article}

\pdfoutput=1
\usepackage[top=3cm,bottom=3cm,left=3cm,right=3cm,marginparwidth=1.75cm]{geometry}

\RequirePackage{algorithm}
\RequirePackage{algorithmic}
\usepackage{natbib}

\renewcommand{\cite}[1]{\citep{#1}}
\setcitestyle{authoryear,round,citesep={;},aysep={,},yysep={;}}

\usepackage{amsfonts}
\usepackage{amsmath}
\usepackage{amssymb}
\usepackage{authblk}
\usepackage{booktabs}
\usepackage[nomessages]{fp}
\usepackage[dvipdfmx]{graphicx}
\usepackage{hyperref}
\usepackage{microtype}
\usepackage{subcaption}
\usepackage{xcolor}

\newcommand{\vv}[1]{\mathbf{#1}}
\newcommand{\R}{\mathbb{R}}
\newcommand{\N}{\mathcal{N}}

\newcommand{\faded}[1] {\FPeval{\result}{max(1-(#1*2.5+0.25),0)}\color[gray]{\result}}

\title{Interpretable VAEs for nonlinear group factor analysis}
\author[1]{Samuel K. Ainsworth}
\author[1]{Nicholas J. Foti}
\author[2]{Adrian KC Lee}
\author[1]{Emily B. Fox}
\affil[1]{School of Computer Science and Engineering, University of Washington}
\affil[2]{Institute for Learning \& Brain Sciences, University of Washington}

\date{February 15, 2018}

\begin{document}

\maketitle

\begin{abstract}
Deep generative models have recently yielded encouraging results in producing subjectively realistic samples of complex data. Far less attention has been paid to making these generative models interpretable. In many scenarios, ranging from scientific applications to finance, the observed variables have a natural grouping.  It is often of interest to understand systems of interaction amongst these groups, and latent factor models (LFMs) are an attractive approach.  However, traditional LFMs are limited by assuming a linear correlation structure.  We present an output interpretable VAE (oi-VAE) for grouped data that models complex, nonlinear latent-to-observed relationships.  We combine a structured VAE comprised of group-specific generators with a sparsity-inducing prior.  We demonstrate that oi-VAE yields meaningful notions of interpretability in the analysis of motion capture and MEG data.  We further show that in these situations, the regularization inherent to oi-VAE can actually lead to improved generalization and learned generative processes.
\end{abstract}

\section{Introduction}
\label{intro}
In many applications there is an inherent notion of groups associated with the observed variables.  For example, in the analysis of neuroimaging data, studies are typically done at the level of regions of interest that aggregate over cortically-localized signals.  In genomics, there are different treatment regimes.  In finance, the data might be described in terms of asset classes (stocks, bonds, \dots) or as collections of regional indices.  Obtaining interpretable and expressive models of the data is critical to the underlying goals of descriptive analyses and decision making.  The challenge arises from the push and pull between interpretability and expressivity in our modeling choices.  Methods for extracting interpretability have focused primarily on linear models, resulting in lower expressivity.  A popular choice in these settings is to consider sparse linear factor models~\citep{Zhao:2016,Carvalho:2008}.  However, it is well known that neural~\cite{Han:2017}, genomic~\cite{Prill:2010}, and financial data~\cite{Harvey:1994}, for example, exhibit complex nonlinearities.  

On the other hand, there has been a significant amount of work on expressive models for complex, high dimensional data.  Building on the framework of latent factor models, the Gaussian process latent variable model (GPLVM)~\cite{Lawrence:2003} introduces nonlinear mappings from latent to observed variables.  A group-structured GPLVM has also been proposed~\cite{Damianou:2012}. However, by relying on GPs, these methods do not scale straightforwardly to large datasets.  In contrast, \textit{deep generative models}~\citep{Kingma:2013,Rezende:2014} have proven wildly successful in efficiently modeling complex observations---such as images---as nonlinear mappings of simple latent representations.  These nonlinear maps are based on deep neural networks and parameterize an observation distribution.  As such, they can viewed as nonlinear extensions of latent factor models.  However, the focus has primarily been on their power as a generative mechanism rather than in the context of traditional latent factor modeling and associated notions of interpretability.

One efficient way of training deep generative models is via the \textit{variational autoencoder} (VAE).  The VAE posits an approximate posterior distribution over latent representations that is parameterized by a deep neural network, called the \textit{inference network}, that maps observations to a distribution over latent variables. This direct mapping of observations to latent variables is called \textit{amortized inference} and alleviates the need to determine individual latent variables for all observations. The parameters of both the generator and inference neural networks can then be determined using Monte Carlo variational inference~\cite{Kingma:2013,Rezende:2014}. The VAE can be interpreted as a nonlinear factor model that provides a scalable means of learning the latent representations.

In this work we propose an \emph{output interpretable VAE} (oi-VAE) for grouped data, where the focus is on interpretable interactions amongst the grouped outputs.  Here, as in standard latent factor models, interactions are induced through shared latent variables.  Interpretability is achieved via sparsity in the latent-to-observed mappings.  To this end, we reformulate the VAE as a nonlinear factor model with a generator neural network for each group and incorporate a sparsity inducing penalty encouraging each latent dimension to influence a small number of correlated groups. We develop an amortized variational inference algorithm for a collapsed variational objective and use a proximal update to learn latent-dimension-to-group interactions. As such, our method scales to massive datasets allowing flexible analysis of data arising in many applications. 

We evaluate the oi-VAE on motion capture and magnetoencephalography datasets.  In these scenarios where there is a natural notion of groupings of observations, we demonstrate the interpretability of the learned features and how these structures of interaction correspond to physically meaningful systems.  Furthermore, in such cases, we show that the regularization employed by oi-VAE leads to better generalization and synthesis capabilities, especially in limited training data scenarios or when the training data might not fully capture the observed space of interest.

\section{Background}
The study of deep generative models is an active area of research in the machine learning community and encompasses probabilistic models of data that can be used to generate observations from the underlying distribution.
The variational autoencoder (VAE)~\cite{Kingma:2013} and the related deep Gaussian model~\cite{Rezende:2014} both propose the idea of amortized inference to perform variational inference in probabilistic models that are parameterized by deep neural networks.  Further details on the VAE specification are provided in Sec.~\ref{sec:model}. The variational objective is optimized with (stochastic) gradient descent and the intractable expectation arising in the objective is evaluated with Monte Carlo samples from the variational distribution. The method is referred to as \emph{Monte Carlo variational inference}, and has become popular for performing variational inference in generative models. See Sec.~\ref{sec:inference}.

The VAE approach has recently been extended to more complex data such as that arising from dynamical systems~\cite{Archer:2015} and also to construct a generative model of cell structure and morphology~\cite{Johnson:Donovan-Maiye:2017}.
Though deep generative models and variational autoencoders have demonstrated the ability to produce convincing samples of complex data from complicated distributions, the learned latent representations are not easily interpretable due to the complex interactions from latent dimensions to the observations, as depicted in Fig.~\ref{fig:bars-data-results}.

A common approach to encourage simple and interpretable models is through use of \textit{sparsity inducing
penalties} such as the \textit{lasso}~\cite{Tibshirani:1994} and \textit{group lasso}~\cite{Yuan:2006}. These methods work by shrinking many model parameters toward zero and have seen great success in regression models, covariance selection~\citep{Danaher:2014}, and linear factor analysis~\cite{Hirose:2012}. The group lasso penalty is of particular interest in our group analysis as it simultaneously shrinks entire groups of model parameters toward zero. Commonly, sparsity inducing penalties are considered in the convex optimization literature due to their computational tractability using proximal gradient descent~\cite{Parikh:2013}.

Though these convex penalties have proven very useful, we cannot apply them directly and obtain a valid generative model.  Instead, we need to consider prior specifications over the parameters of the generator network that likewise yield sparsity. Originally, the Bayesian approach to sparsity was based on the spike-and-slab prior, a two-component mixture that puts some probability on a model parameter being exactly zero (the spike) and some probability of the parameter taking on non-zero values (the slab)~\cite{Mitchell:1988}. Unfortunately, inference in models with the spike-and-slab prior is difficult because of the combinatorial nature of the resulting posterior.

% Unfortunately, the spike-and-slab prior is difficult to perform inference with because of the combinatorial nature of the resulting posterior.

Recently, Bayesian formulations of sparsity inducing penalties take the form of hierarchical prior distributions that shrink many model parameters to small values (though not exactly zero).
Such \emph{global-local shrinkage} priors encapsulate a wide variety of hierarchical Bayesian priors that attempt to infer interpretable models, such as the horseshoe prior~\cite{Bhadra:2016}. These priors also result in efficient inference algorithms.

%A Bayesian group-lasso that is marginally equivalent to the convex group-lasso penalty has recently been proposed~\cite{}.
A sophisticated hierarchical Bayesian prior for sparse group linear factor analysis has recently been developed by~\cite{Zhao:2016}. This prior encourages both a sparse set of factors to be used as well as having the factors themselves be sparse. Additionally, the prior admits an efficient inference scheme via expectation-maximization.
Sparsity inducing hierarchical Bayesian priors have been applied to Bayesian deep learning models to learn the complexity of the deep neural network~\cite{Louizos:2017,Ghosh:2017}. Our focus, however, is on using (structured) sparsity-inducing hierarchical Bayesian priors in the context of deep learning for the sake of interpretability, as in linear factor analysis, rather than model selection.

\begin{figure}
\centering
\begin{subfigure}{0.45\columnwidth}
  \centering
  \includegraphics[scale=0.75]{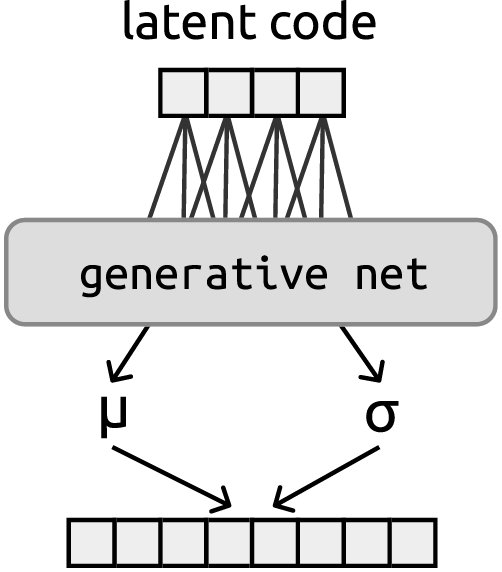}
  % \caption{Variational autoencoder}
  \label{fig:vae-diagram}
\end{subfigure}
\begin{subfigure}{0.45\columnwidth}
  \centering
  \includegraphics[scale=0.75]{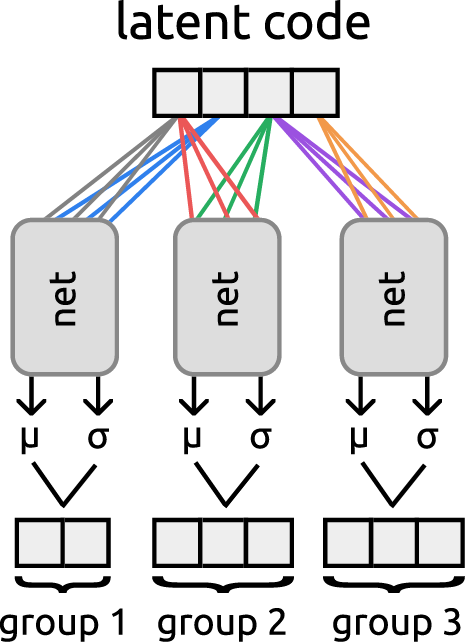}
%  \caption{Output interpretable VAE}
  \label{fig:oi-vae-diagram}
\end{subfigure}
\caption{VAE (\emph{left}) and oi-VAE (\emph{right}) generative models. The oi-VAE considers group-specific generators and a linear latent-to-generator mapping with weights from a single latent dimension to a specific group sharing the same color. The group-sparse prior is applied over these grouped weights.}
\label{fig:model_diagram}
\end{figure}

\section{The OI-VAE model}
\label{sec:model}

We frame our proposed output interpretable VAE (oi-VAE) method using the same terminology as the VAE. Let $\vv{x} \in \R^{D}$ denote a $D$-dimensional observation and $\vv{z} \in \R^{K}$ denote the associated $K$-dimensional latent representation. We then write the generative process of the model as:
\begin{align}
\label{eq:vae_factor_model}
	\vv{z} &\sim \N(\vv{0}, \vv{I}) \\
	\vv{x} &\sim \N(f_\theta (\vv{z}), \vv{D}),
\end{align}
where $\vv{D}$ is a diagonal matrix containing the marginal variances of each component of $\vv{x}$. The generator is encoded with the function $f_\theta(\cdot)$ specified as a deep neural network with parameters $\theta$.
Note that the formulation in Eq.~\eqref{eq:vae_factor_model} is simpler than that described in \citet{Kingma:2013} as we assume the observation variances are global parameters and not observation specific.  This simplifying assumption follows from that of traditional factor models, but could easily be relaxed.

%In the standard VAE model, samples $\textbf{x}$ are drawn by first taking some random latent code $\textbf{z}$ sampled from a diagonal normal distribution and then feeding it through a generative network to produce the parameters of a likelihood distribution, usually normal or Bernoulli:
%\begin{align}
%\textbf{z} &\sim \mathcal{N}(\mathbf{0}, \mathbf{I}) \\
%\textbf{x} &\sim \rho (f_\theta (\textbf{z}))
%\end{align}
%where $f_\theta (\cdot)$ is the generative net with parameters $\theta$ and $\rho$ denotes the likelihood distribution density.
%This is all fine, but the trick comes when doing inference in this model. We would like to maximize the probability of our observed $\mathbf{x}$'s by integrating out all possible $\mathbf{z}$'s. It should not be surprising that doing so is not analytically or computationally tractable. Instead (Kingma and Welling) \cite{} apply a variational approximation, $q_\phi(\mathbf{z} | \mathbf{x})$ parameterized by $\phi$, which we hope to optimize towards $q_\phi(\mathbf{z} | \mathbf{x}) \approx p(\mathbf{z} | \mathbf{x})$.

When our observations $\vv{x}$ admit a natural grouping over the components, we write $\vv{x}$ as $[\vv{x}^{(1)},\ldots,\vv{x}^{(G)}]$ for our $G$ groups.  We model the components within each group $g$ with separate generative networks $f^{(g)}_{\theta_g}$ parameterized by $\theta_g$. It is possible to share generator parameters $\theta_g$ across groups, however we choose to model each separately.  Critically, the latent representation $\vv{z}$ is shared over all the group-specific generators.  In particular:
\begin{align}
\vv{z} &\sim \mathcal{N}(\vv{0}, \vv{I}) \\
\vv{x}^{(g)} &\sim \N(f^{(g)}_{\theta_g}(\vv{z}), \vv{D}_g).
\end{align}
To this point, our specified group-structured VAE can leverage within-group correlation structure and between-group independencies.  However, one of the main goals of this framework is to capture interpretable relationships between group-specific activations through the latent representation.  Note that it is straightforward to apply different likelihoods on different groups, although we did not have reason to do so in our experiments.

Inspired by the sparse factor analysis literature, we extract notions of interpretable interactions through inducing sparse latent-to-group mappings.  Specifically, we insert a group-specific linear transformation $\vv{W}^{(g)} \in \R^{p\times K}$ between the latent representation $\vv{z}$ and the group generator $f^{(g)}$:
\begin{align}
\vv{x}^{(g)} &\sim \N(f^{(g)}_\theta (\mathbf{W}^{(g)} \textbf{z}),\vv{D}_g).
\end{align}
We refer to $\vv{W}^{(g)}$ as the \textit{latent-to-group matrix}.  We assume that the input dimension $p$ per generator is the same, but this could be generalized.  When the $j$th column of the group-$g$ latent-to-group matrix, $\vv{W}_{:,j}^{(g)}$, is all zeros then the $j$th latent dimension, $\vv{z}_j$, will have no influence on group $g$.  To induce this column-wise sparsity, we place a hierarchical Bayesian prior on the columns $\vv{W}_{:,j}^{(g)}$ as follows \cite{Kyung:2010}:
\begin{align}
\gamma_{gj}^2 &\sim \text{Gamma}\left(\frac{p + 1}{2}, \frac{\lambda^2}{2}\right) \\
\vv{W}^{(g)}_{\boldsymbol{\cdot}, j} &\sim \mathcal{N}(\vv{0}, \gamma_{gj}^2 \vv{I})
\end{align}
where Gamma($\cdot,\cdot$) is defined by shape and rate, and $p$ denotes the number of rows in each $\vv{W}^{(g)}$. The rate parameter $\lambda$ defines the amount of sparsity, with larger $\lambda$ implying more column-wise sparsity in $\vv{W}^{(g)}$. Marginalizing over $\gamma_{gj}^2$ induces group sparsity over the columns of $\vv{W}^{(g)}$; the MAP of the resulting posterior is equivalent to a group lasso penalized objective~\cite{Kyung:2010}. 

While we are close to a workable model, one wrinkle remains. Unlike linear factor models, the deep structure of our model permits it to push rescaling across layer boundaries without affecting the end behavior of the network. In particular, it is possible---and in fact encouraged behavior---to learn a set of $\vv{W}^{(g)}$ matrices with very small weights only to have the values revived to ``appropriate'' magnitudes in the following layers of $f^{(g)}_{\theta_g}$. In order to mitigate such behavior we additionally place a standard normal prior on the parameters of each generative network, $\theta_g \sim \mathcal{N}(\vv{0}, \vv{I})$, completing the model specification.

\paragraph{Special cases of the oi-VAE}
There are a few notable special cases of our oi-VAE framework. When we treat the observations as forming a single group, the model resembles a traditional VAE since there is a single generator.  However, the sparsity inducing prior still has an effect that differs from the standard VAE specification.  In particular, by shrinking columns of $\vv{W}$ the prior will essentially encourage a sparse subset of the components of $\vv{z}$ to be used to explain the data, similar to a traditional sparse factor model. Note that the $\vv{z}$'s themselves will not necessarily be sparse, but the columns of $\vv{W}$ will indicate which components are used. (Here, we drop the $g$ superscript on $\vv{W}$.) This regularization can be advantageous to apply even when the data only has one group as it can provide improved generalization performance in the case of limited training data.
Another special case arises when the generator networks are given by the identity mapping.  In this case, the only transformation of the latent representation is given by $\vv{W}^{(g)}$ and the oi-VAE reduces to a group sparse linear factor model.  %If the number of groups is one, then we obtain standard sparse linear factor model.

%\paragraph{Relative to the VAE}  The oi-VAE provides many potential advantages over the standard VAE formulation.  One is in terms of interpretability, as further discussed in Sec.~\ref{sec:interpretability}.  Another is through the regularization introduced by the sparsity-inducing prior.  As we demonstrate in the experiments of Sec.~\ref{sec:mocap}, this regularization not only leads to interpretability, but can also improve test log-likelihood in the case of limited training data.  On the other hand, the decomposition into group-specific generators implies that \ebf{finish thought}.
\section{Interpretability of the oi-VAE}
\label{sec:interpretability}

In the oi-VAE, each latent factor influences a sparse set of the observational groups. The interpretability garnered from this sparse structure is two-fold:

\paragraph{Disentanglement of latent embeddings} By associating each component of $\vv{z}$ with only a sparse subset of the observational groups, we are able to quickly identify \emph{disentangled} representations in the latent space.  That is, by penalizing interactions between the components of $\mathbf{z}$ and each of the groups, we effectively force the model to arrive at a representation that minimizes correlation across the components of $\mathbf{z}$, encouraging each dimension to capture distinct modes of variation.  For example, in Table~\ref{table:mocap-topic-model-thing} we see that each of the dimensions of the latent space learned on motion capture recordings of human motion corresponds to a direction of variation relevant to only a subset of the joints (groups) that are used in specific submotions related to walking. Additionally, it is observed that although the VAE and oi-VAE have similar reconstruction performance the meaningfully disentangled latent representation allows oi-VAE to produce superior unconditional random samples.

\paragraph{Discovery of group interactions} From the perspective of the latent representation $\vv{z}$, each latent dimension influences only a sparse subset of the observational groups.  As such, we can view the observational groups associated with a specific latent dimension as a related system of sorts.  For example, in neuroscience the groups could correspond to different brain regions from a standard parcellation. If a particular dimension of $\vv{z}$ influences the generators of a small set of groups, then those groups can be interpreted as a system of regions that can be treated as a unit of analysis. Such an approach is attractive in the context of analyzing functional connectivity from MEG data where we seek modules of highly correlated regions. See the experiments of Sec.~\ref{sec:MEG}. Likewise, in our motion capture experiments of Sec.~\ref{sec:mocap}, we see (again from Table~\ref{table:mocap-topic-model-thing}) how we can treat collections of joints as a system that covary in meaningful ways within a given human motion category.  

\mbox{}\\
Broadly speaking, the relationship between dimensions of $\vv{z}$ and observational groups can be thought of as a bipartite graph in which we can quickly identify correlation and independence relationships among the groups themselves.  The ability to expose or refute correlations among observational groups is attractive as an exploratory scientific tool independent of building a generative model.  This is especially useful since standard measures of correlation are linear, leaving much to be desired in the face of high-dimensional data with many potential nonlinear relationships. Our hope is that oi-VAE serves as one initial tool to spark a new wave of interest in nonlinear factor models and their application to complicated and rich data across a variety of fields.

% For example, a notion of interpretablility arises from the perspective of the observational groups.  Each group is comprised of observed variations attributed to variations in a subset of the latent dimensions.  For example, a specific joint (e.g., left femur) can be directly linked to its relevant modes of variation.  See Fig.~\ref{fig:mocap-connectivity-matrices}.  One can think of this as a feature model where each group chooses a subset of the latent dimensions relevant to its observations.

It is worth emphasizing that the goal is \textit{not} to learn sparse representations in the $\mathbf{z}$'s. Sparsity in $\mathbf{z}$ may be desirable in certain contexts, but it does not actually provide any interpretability in the data generating process. In fact, excessive compression of the latent representation $\vv{z}$ through sparsity could be detrimental to interpretability.

% Finally, note that the VAE does not afford any ability to the interpret these notions of latent-to-observed interactions.

%Since the relationship between dimensions of $\vv{z}$ and observational groups can be thought of as a bipartite graph, we can also quickly identify correlation and independence relationships among the groups themselves. The disentanglement of latent embeddings is an attractive feature of a generative modeling framework, but the ability to expose or refute correlations among observational groups is attractive as an exploratory scientific tool independent of building a generative model. This is especially attractive since standard measures of correlation are linear, leaving much to be desired in the face of high-dimensional data with many potential nonlinear relationships. {\color{red} This can be helpful in such and such example.}

\section{Collapsed variational inference}
\label{sec:inference}
Traditionally, VAEs are learned by applying stochastic gradient methods directly to the evidence lower bound (ELBO):
\begin{align*}
\log p(\vv{x}) \geq \mathbb{E}_{q_{\phi}(\vv{z}|\vv{x})}[\log p_\theta(\vv{x}, \vv{z}) - \log q_\phi(\vv{z}|\vv{x})],
\end{align*}
where $q_\phi(\vv{z}|\vv{x})$ denotes the amortized posterior distribution of $\vv{z}$ given observation $\vv{x}$, parameterized with a deep neural network with weights $\phi$. Using a neural network to parameterize the observation distribution $p(\vv{x}|\vv{z})$ as in Eq.~\eqref{eq:vae_factor_model} makes the expectation in the ELBO intractable. To address this, the VAE employs Monte Carlo variational inference (MCVI)~\cite{Kingma:2013}: The troublesome expectation is approximated with samples of the latent variables from the variational distribution, $\vv{z} \sim q_\phi(\vv{z}|\vv{x})$, where $q_\phi(\vv{z}|\vv{x})$ is \textit{reparameterized} to allow differentiating through the expectation operator and reduce gradient variance. %This procedure is commonly known as Monte Carlo variational inference (MCVI)~\cite{}.

We extend the basic VAE amortized inference procedure to incorporate our sparsity inducing prior over the columns of the latent-to-group matrices. The naive approach of optimizing variational distributions for the $\gamma_{gj}^2$ and $\vv{W}_{\cdot,j}^{(g)}$ will not result in true sparsity of the columns $\vv{W}_{\cdot,j}^{(g)}$. Instead, we consider a collapsed variational objective function. Since our sparsity inducing prior over $\vv{W}_{\cdot,j}^{(g)}$ is marginally equivalent to the convex group lasso penalty we can use proximal gradient descent on the collapsed objective and obtain true group sparsity~\cite{Parikh:2013}.
Following the standard VAE approach of \citet{Kingma:2013}, we use simple point estimates for the variational distributions on the neural network parameters $\mathcal{W} = \left(\mathbf{W}^{(1)}, \cdots, \mathbf{W}^{(G)}\right)$ and $\theta = \left(\theta_1,\ldots,\theta_G\right)$. %Doing so allows us to optimize directly over the generative network parameters as opposed to samples of network parameters.
We take $q_\phi(\vv{z}|\vv{x}) = \N(\mu(\vv{x}), \sigma^2(\vv{x})))$ where the mean and variances are parameterized by an inference network with parameters $\phi$.

%However, the introduction of sparsity-inducing priors brings new complexity to the optimization problem, both in terms of rate of convergence as well as the resulting sparsity level. For example it is well known that subgradient methods perform poorly on many sparsity-inducing objectives \cite{}. Standard variational inference or MCMC in a fully Bayesian sparsity model may avoid subgradients but introduces new variables and converges slowly. Here we outline an approach to learning OI-VAEs that carefully exploits their structure to build a collapsed variational objective which in turn can be efficiently optimized with proximal gradient descent.

\subsection{The collapsed objective}
We construct a collapsed variational objective by marginalizing the $\gamma_{gj}^2$ to compute $\log p(\vv{x})$ as:  
\begin{equation*}
\begin{split}
\log\ p(\mathbf{x}) &= \log \int p(\mathbf{x} | \mathbf{z}, \mathcal{W}, \theta) p(\mathbf{z}) p(\mathcal{W} | \gamma^2) p(\gamma^2) p(\theta) \,d\gamma^2 \,dz \\
% Typesetting this line to be within the column boundaries is a bit of a challenge.
% &= \log \int \left( \int p(\mathcal{W} | \gamma^2) p(\gamma^2) \,d\gamma^2 \right) \frac{p(x | z, \mathcal{W}, \theta) p(z) p(\theta)}{q_\phi(z)} q_\phi(z) \,dz \\
% &= \log \int \left( \int p(\mathcal{W}, \gamma^2) \,d\gamma^2 \right) \frac{p(x | z, \mathcal{W}, \theta) p(z) p(\theta)}{q_\phi(z)} q_\phi(z) \,dz \\
&= \log \int \left( \int p(\mathcal{W}, \gamma^2) \,d\gamma^2 \right) \frac{p(\mathbf{x} | \mathbf{z}, \mathcal{W}, \theta) p(\mathbf{z}) p(\theta)}{q_\phi(\mathbf{z} | \mathbf{x}) / q_\phi(\mathbf{z} | \mathbf{x})} \,dz \\
% &= \log \int \frac{p(x | z, \mathcal{W}, \theta) p(z) p(\mathcal{W}) p(\theta)}{q_\phi(z)} q_\phi(z) \,dz \\
&\geq \mathbb{E}_{q_\phi(\mathbf{z} | \mathbf{x})} \left[ \log p(x | \mathbf{z}, \mathcal{W}, \theta) \right]  - \mathbb{KL}(q_\phi(\mathbf{z} | \mathbf{x}) || p(\mathbf{z})) \\
&\qquad + \log p(\theta) - \lambda \sum_{g,j} || \mathbf{W}^{(g)}_{\boldsymbol{\cdot}, j} ||_2 \\
&\triangleq \mathcal{L}(\phi, \theta, \mathcal{W}).
\end{split}
\end{equation*}
%where $\mathcal{W}$ denotes the set of all $\vv{W}^{(g)}$ for all groups.
Importantly, the columns of the latent-to-group matrices $\vv{W}_{\cdot,j}^{(g)}$ appear in a 2-norm penalty in the new collapsed ELBO. This is exactly a group lasso penalty on the columns of $\vv{W}_{\cdot,j}^{(g)}$ and encourages the entire vector to be set to zero.

Now our goal becomes maximizing this collapsed ELBO over $\phi, \theta, \mathcal{W}$.
%\begin{equation}
%\label{eq:elbo_opt}
%\max_{\phi, \theta, \mathcal{W}} \mathcal{L}(\phi, \theta, \mathcal{W}).
%% = \hat{\mathcal{L}}(\phi, \theta, \mathcal{W})- \lambda \sum_{g,j} || \mathbf{W}^{(g)}_{\boldsymbol{\cdot}, j} ||_2.
%\end{equation}
% And immediately we arrive at one benefit of collapsing the $\gamma_{gj}^2$ scale parameters: we are left with an objective akin to standard group lasso. Within this regime it becomes possible to leverage efficient proximal gradient descent updates on $\mathcal{W}$ in order to speed up convergence. On the remaining parameters we still have the flexibility to use any other off-the-shelf optimization method of our choice such as RMSProp~\cite{} or Adam~\cite{}. 
%
Since this objective contains a standard group lasso penalty, we can leverage efficient proximal gradient descent updates on the latent-to-group matrices $\mathcal{W}$ as detailed in Sec.~\ref{sec:proxgrad}. Proximal algorithms achieve better rates of convergence than sub-gradient methods and have shown great success in solving convex objectives with group lasso penalties.
We can use any off-the-shelf optimization method for the remaining neural net parameters, $\theta_g$ and $\phi$.  %We choose to use Adam~\cite{Kingma:2014}.

%One extremely convenient consequence of using a collapsing the $\gamma_{gj}^2$ scale parameters is that it leads us to an objective akin to standard group lasso. Within this regime it becomes possible to leverage efficient proximal gradient descent updates on $\mathcal{W}$ in order to speed up convergence. On the remaining parameters we still have the flexibility to use any other off-the-shelf optimization method of our choice such as RMSProp~\cite{} or Adam~\cite{}. 

\subsection{Proximal gradient descent}
\label{sec:proxgrad}

Proximal gradient descent algorithms are a broad class of optimization techniques for separable objectives with both differentiable and potentially non-differentiable components,
\begin{align}
\min_{x} g(x) + h(x),
\end{align}
where $g(x)$ is differentiable and $h(x)$ is potentially non-smooth or non-differentiable~\cite{Parikh:2013}. Stochastic proximal algorithms are well-studied for convex optimization problems. Recent work has shown that they are guaranteed to converge to a local stationary point even if the objective is comprised of a non-convex $g(x)$ as long as the non-smooth $h(x)$ is convex~\cite{Reddi:2016}.
The usual tactic is to take gradient steps on $g(x)$ followed by ``corrective'' \textit{proximal} steps to respect $h(x)$:
\begin{align}
\label{eq:proximal_step}
x_{t+1} = \text{prox}_{\eta h} (x_t - \eta \nabla g(x_t))
\end{align}
where $\text{prox}_f (x)$ is the proximal operator for the function $f$. For example, if $h(x)$ is the indicator function for a convex set then the proximal operator is simply the projection operator onto the set and the update in Eq.~\eqref{eq:proximal_step} is projected gradient.
% where the $\text{prox}_f (x)$ operator is defined as
% \begin{align}
% \text{prox}_f (x) \triangleq \argmin_v \left( f(v) + \frac{1}{2} ||x - v||_2^2 \right).
% \end{align}
Expanding the definition of $\text{prox}_{\eta h}$ in Eq.~\eqref{eq:proximal_step}, one can show that the proximal step corresponds to minimizing $h(x)$ plus a quadratic approximation to $g(x)$ centered on $x_t$. For $h(x) = \lambda ||x||_2$, the proximal operator is given by
\begin{align}
\text{prox}_{\eta h} (x) = \frac{x}{||x||_2} \left(||x||_2 - \eta \lambda \right)_+ 
\end{align}
where $(v)_+ \triangleq \max (0, v)$~\cite{Parikh:2013}. This operator is especially convenient since it is both cheap to compute and results in machine-precision zeros, unlike many hierarchical Bayesian approaches to sparsity that result in small but non-zero values. These methods require an extra thresholding step that our oi-VAE method does not due to attain exact zeros. Geometrically, this operator reduces the norm of $x$ by $\eta \lambda$, and shrinks $x$'s with $||x||_2 \leq \eta \lambda$ to zero.

% One extremely convenient consequence of using a collapsed variational inference is that it leads us to an objective akin to standard group lasso. Within this regime it becomes possible to leverage efficient proximal gradient descent updates on $\mathcal{W}$ in order to speed up convergence. On the remaining parameters we still have the flexibility to use any other off-the-shelf optimization method of our choice such as RMSProp~\cite{} or Adam~\cite{}. 

We experimented with standard (non-collapsed) variational inference as well as other schemes, but found that collapsed variational inference with proximal updates provided faster convergence and succeeded in identifying sparser models than other techniques. In practice we apply proximal stochastic gradient updates per Eq.~\eqref{eq:proximal_step} on the $\mathcal{W}$ matrices and Adam~\cite{Kingma:2014} on the remaining parameters. See Alg.~\ref{alg:vi} for oi-VAE pseudocode.

\begin{algorithm}[tb]
   \caption{Collapsed VI for oi-VAE}
   \label{alg:vi}
\begin{algorithmic}
   \STATE {\bfseries Input:} data $\mathbf{x}^{(i)}$, sparsity parameter $\lambda$
   \STATE Let $\tilde{\mathcal{L}}$ be $\mathcal{L}(\phi, \theta, \mathcal{W})$ but without $- \lambda \sum_{g,j} || \mathbf{W}^{(g)}_{\boldsymbol{\cdot}, j} ||_2$.
   \REPEAT
   \STATE Calculate $\nabla_\phi \tilde{\mathcal{L}}$, $\nabla_\theta \tilde{\mathcal{L}}$, and $\nabla_\mathcal{W} \tilde{\mathcal{L}}$.
   \STATE Update $\phi$ and $\theta$ with an optimizer of your choice.
   \STATE Let $\mathcal{W}_{t+1} = \mathcal{W}_{t} - \eta \nabla_\mathcal{W} \tilde{\mathcal{L}}$.
   \FORALL{groups $g$, $j=1$ {\bfseries to} $K$}
       \STATE Set $\mathbf{W}^{(g)}_{\boldsymbol{\cdot}, j} \gets \frac{\mathbf{W}^{(g)}_{\boldsymbol{\cdot}, j}}{||\mathbf{W}^{(g)}_{\boldsymbol{\cdot}, j}||_2} \left(||\mathbf{W}^{(g)}_{\boldsymbol{\cdot}, j}||_2 - \eta \lambda \right)_+ $
   \ENDFOR
   \UNTIL{convergence in both $\hat{\mathcal{L}}$ and $- \lambda \sum_{g,j} || \mathbf{W}^{(g)}_{\boldsymbol{\cdot}, j} ||_2$}
\end{algorithmic}
\end{algorithm}

%{ \color{red} 
%Explain why proximal gradient descent is nice. This is sort of explained. Have we really sold it enough though?
%}

\section{Experiments}
\label{sec:experiments}
\subsection{Synthetic data}
\begin{figure}[t!]
\begin{tabular}{cccc}
\includegraphics[width=0.24\columnwidth]{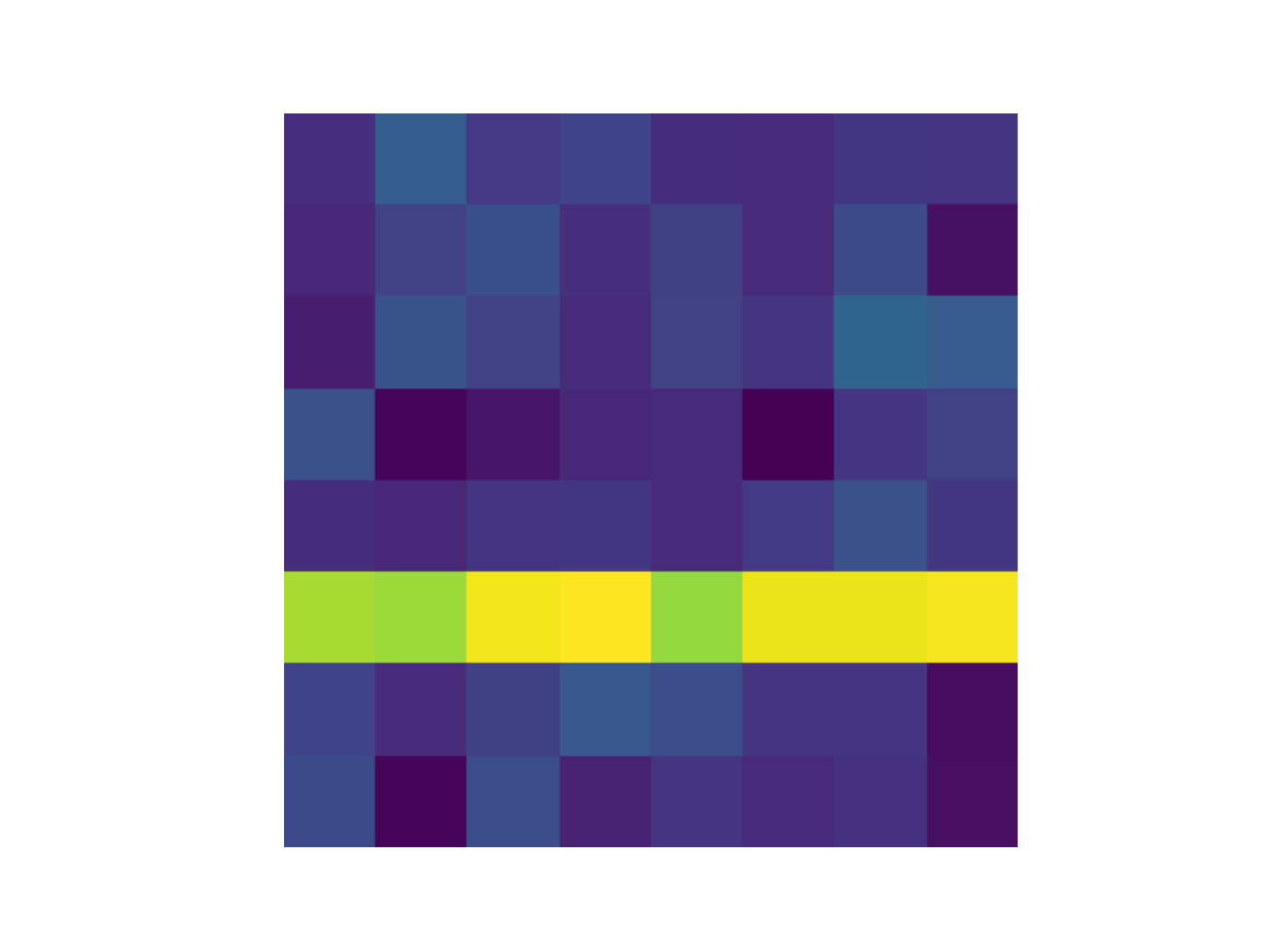} &\hspace{-0.2in} \includegraphics[width=0.24\columnwidth]{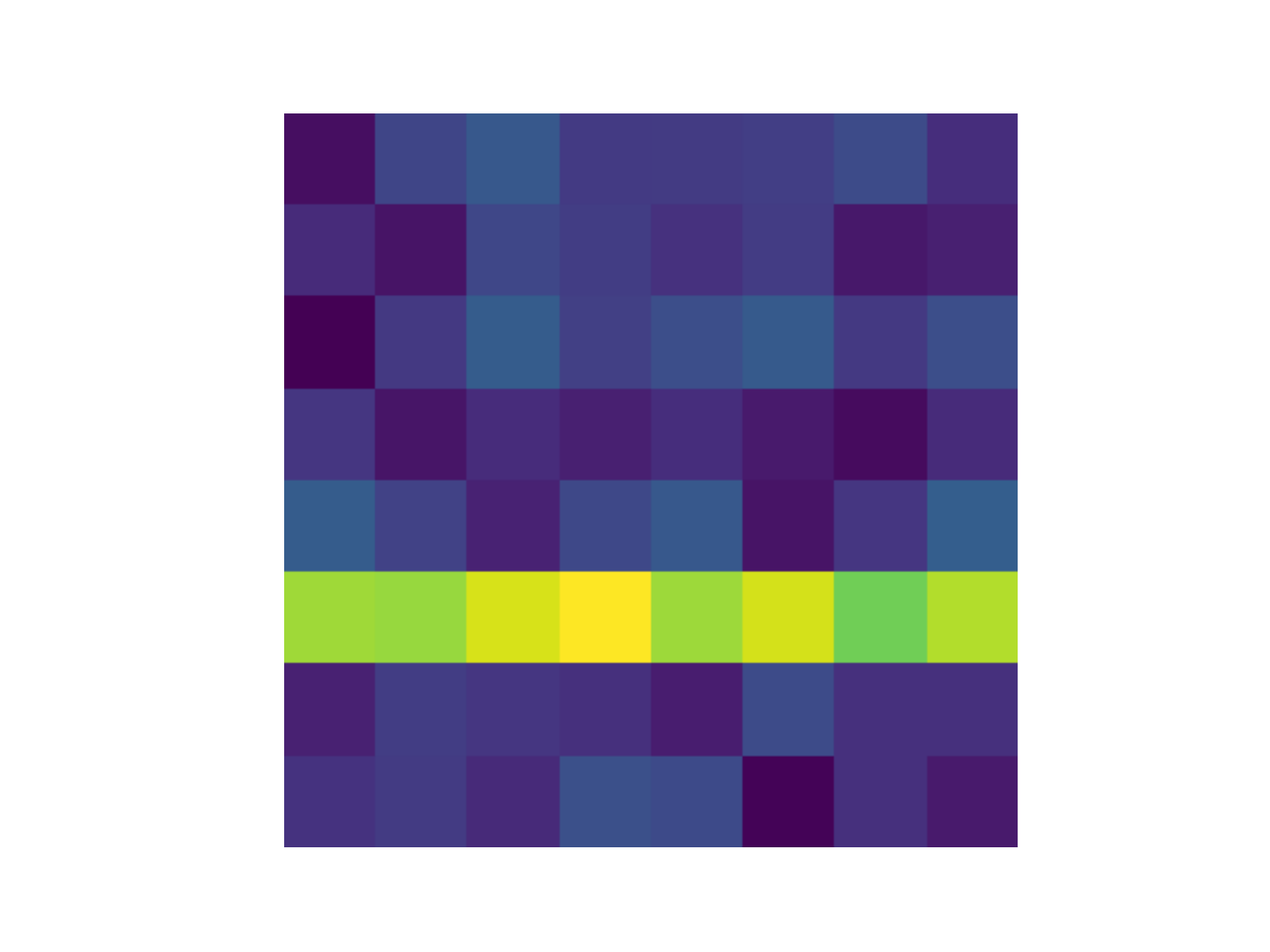} &\hspace{-0.2in}  \includegraphics[width=0.24\columnwidth]{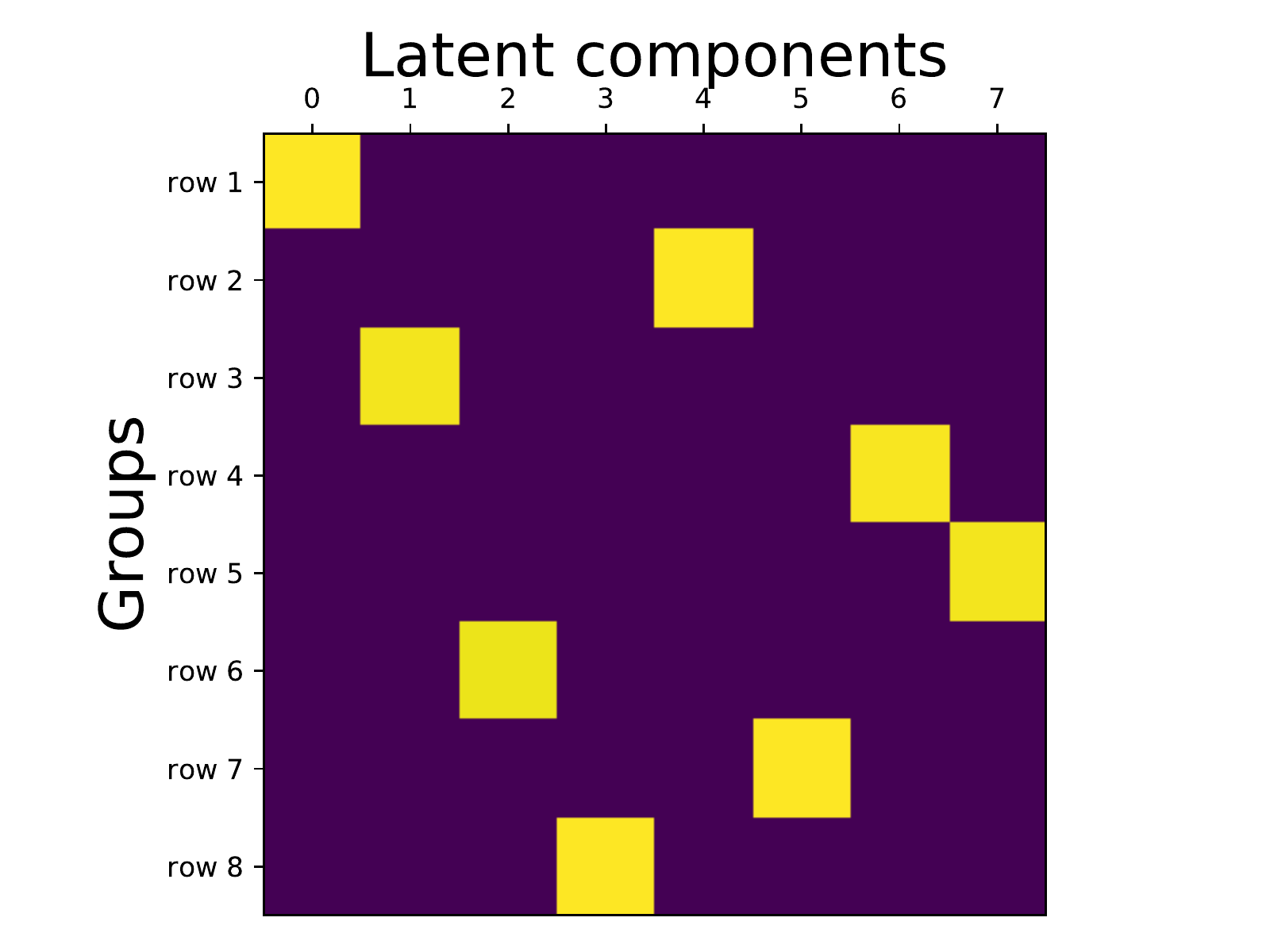} &\hspace{-0.2in}  \includegraphics[width=0.24\columnwidth]{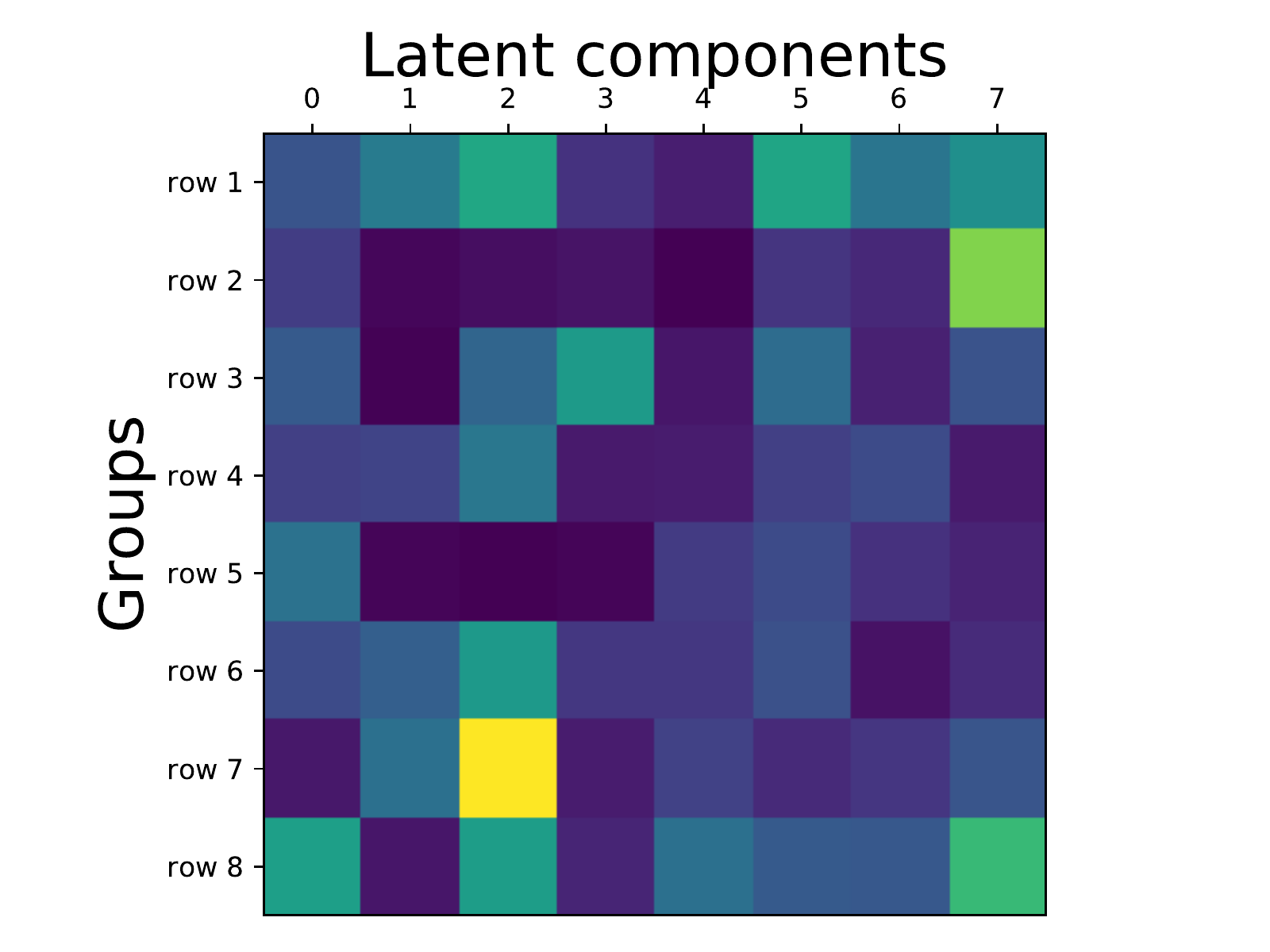}\\
(a) &\hspace{-0.2in}(b) &\hspace{-0.2in} (c) &\hspace{-0.2in} (d)
\end{tabular}
\caption{oi-VAE results on synthetic bars data. (a) Example image and (b) oi-VAE reconstruction. Learned oi-VAE $\vv{W}_{\cdot,j}^{(g)}$ for (c) $\lambda=1$ and (d) $\lambda=0$ (group structure, but no sparsity). In this case, training and test error numbers are nearly identical.}
\label{fig:bars-data-results}
\end{figure}

In order to evaluate oi-VAE's ability to identify sparse models on well-understood data, we generated $8 \times 8$ images with one randomly selected row of pixels shaded and additive noise corrupting the entire image. We then built and trained an oi-VAE on the images with each group defined as an entire row of pixels in the image. We used an $8$-dimensional latent space in order to encourage the model to associate each dimension of $\vv{z}$ with a unique row in the image. Results are shown in Fig.~\ref{fig:bars-data-results}. Our oi-VAE successfully disentangles each of the dimensions of $\vv{z}$ to correspond to exactly one row (group) of the image. % and correctly identified the standard deviation of the additive noise.
We also trained an oi-VAE with a $16$-dimensional latent space (see the Supplement) and see that when additional latent components are not needed to describe any group they are pruned from the model.

% we can put this back in if we rerun the experiment and think it's interesting.
%Although the primary goal of OI-VAE is not to compete with VAEs in terms of reconstruction performance, we also evaluated its performance on held out test data in the small training data regime (eg. less than 100 samples) and found that it outperformed standard VAEs, successfully achieving essentially zero test error with as few as 32 training samples indicating that our sparsity-inducing prior can in fact lead to improved performance.
%{\color{red} TODO this was done with the standard autoencoder and without noise IIRC. We should re-run in the new model.}

\subsection{Motion Capture}
\label{sec:mocap}
Using data collected from CMU's motion capture database we evaluated oi-VAE's ability to handle complex physical constraints and interactions across groups of joint angles while simultaneously identifying a sparse decomposition of human motion. The dataset consists of 11 examples of \texttt{walking} and one example of \texttt{brisk walking} from the same subject.  The recordings measure 59 joint angles split across 29 distinct joints.  The joint angles were normalized from their full ranges to lie between zero and one.  We treat the set of measurements from each distinct joint as a group; since each joint has anywhere from 1 to 3 observed degrees of freedom, this setting demonstrates how oi-VAE can handle variable-sized groups.  For training, we randomly sample 1 to 10 examples of \texttt{walking}, resulting in up to 3791 frames.  Our experiments evaluate the following performance metrics: interpretability of the learned interaction structure amongst groups and of the latent representation; test log-likelihood, assessing the model's generalization ability; and both conditional and unconditional samples to evaluate the quality of the learned generative process.  In all experiments, we use $\lambda=1$ with the reconstruction loss normalized by the dataset size. For further details on the specification of all considered models (VAE and oi-VAE), see the Supplement.

\begin{figure}[t!]
\centering
\centerline{\includegraphics[width=0.7\columnwidth]{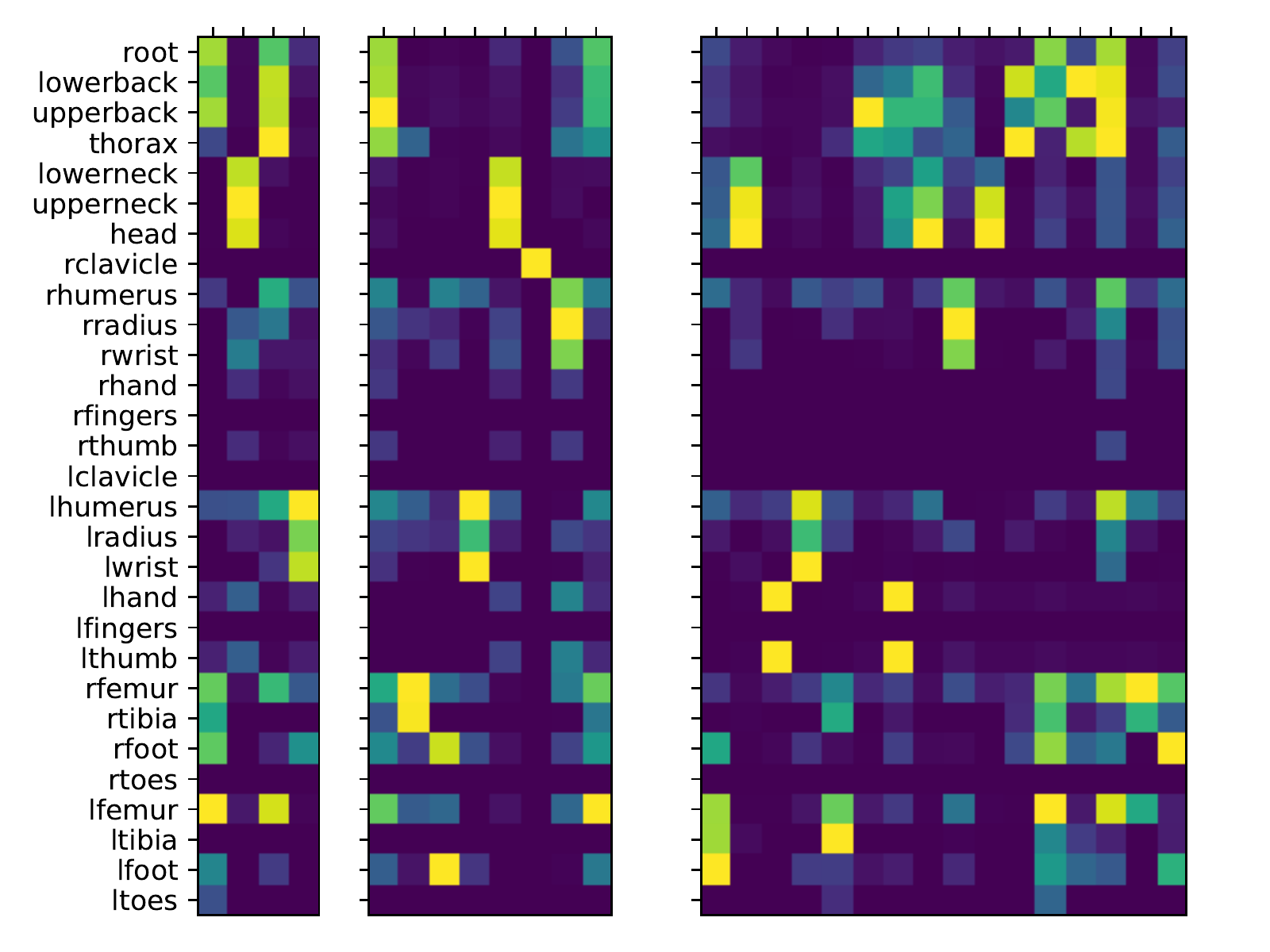}}
\caption{\small oi-VAE results on motion capture data with $K=4$, $8$, and $16$. Rows correspond to group generators for each of the joints in the skeleton, columns correspond to individual dimensions of the latent code, and values in the heatmap show the strength of the latent-to-group mappings $\vv{W}_{\cdot,j}^{(g)}$. Note, joints that experience little motion when walking---clavicles, fingers, and toes---have been effectively pruned from the latent code in all 3 models.}
\label{fig:mocap-connectivity-matrices}
\end{figure}

To begin, we train our oi-VAE on the full set of 10 training trials with the goal of examining the learned latent-to-group mappings.  To explore how the learned disentangled latent representation varies with latent dimension $K$, we use $K=4$, $8$, and $16$.  The results are summarized in Fig.~\ref{fig:mocap-connectivity-matrices}.  We see that as $K$ increases, individual ``features'' (i.e., components of $\mathbf{z}$) are refined to capture more localized anatomical structures.  For example, feature 2 in the $K=4$ case turns into feature 7 in the $K=16$ case, but in that case we also add feature 3 to capture just variations of \texttt{lfingers}, \texttt{lthumb} separate from \texttt{head}, \texttt{upperneck}, \texttt{lowerneck}. Likewise, feature 2 when $K=16$ represents \texttt{head}, \texttt{upperneck}, \texttt{lowerneck} separately from \texttt{lfingers}, \texttt{lthumb}. To help interpret the learned disentangled latent representation, for the $K=16$ embedding we provide lists of the 3 joints per dimension that are most strongly influenced by that component.  From these lists, we see how the learned decomposition of the latent representation has an intuitive anatomical interpretation.  For example, in addition to the features described above, one of the very prominent features is feature 14, which jointly influences the \texttt{thorax}, \texttt{upperback}, and \texttt{lowerback}.  Collectively, these results clearly demonstrate how the oi-VAE provides meaningful interpretability.  We emphasize that it is not even possible to make these types of images or lists for the VAE.

\begin{figure}[t]
\centering
\includegraphics[width=0.95\columnwidth]{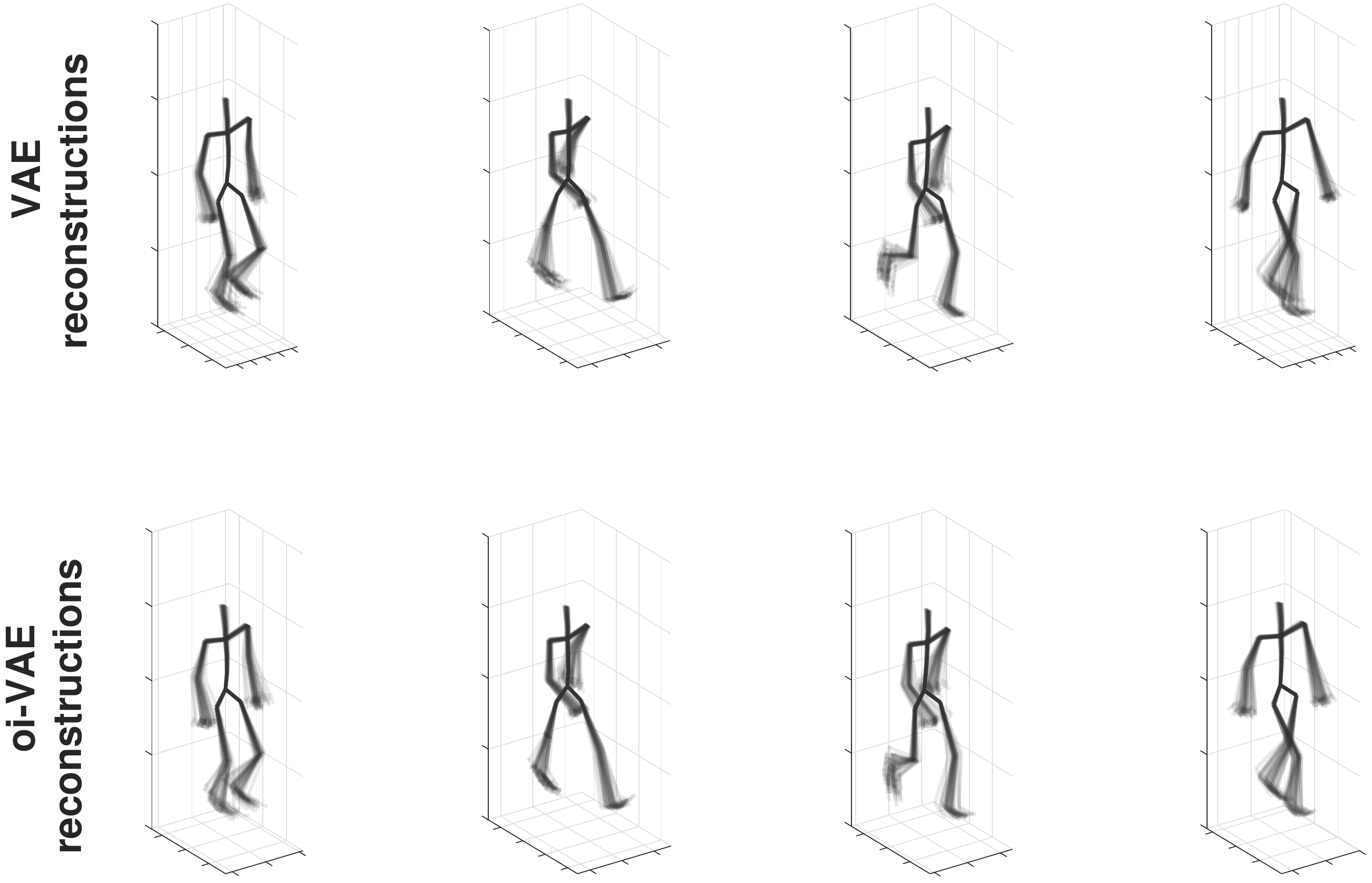}
\caption{\small Samples an oi-VAE model trained on walking data and conditioned on an out-of-sample video frame. We can see that oi-VAE has learned noise patterns that reflect the gait as opposed to arbitrary perturbative noise.}
\label{fig:mocap-reconstruction-samples}
\end{figure}

\begin{figure}[t]
\centering
\includegraphics[width=0.95\columnwidth]{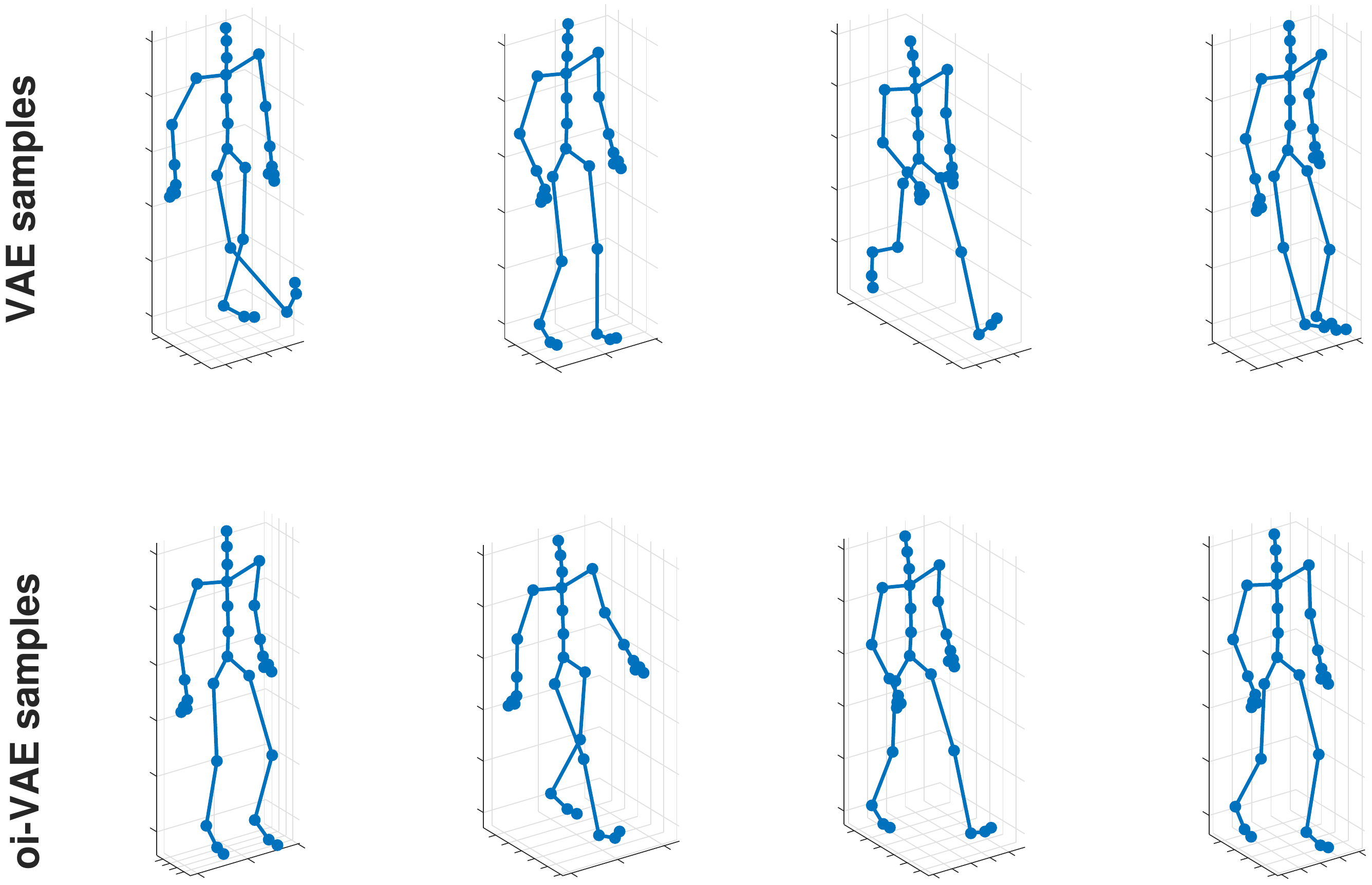}
\caption{\small Representative unconditional samples from oi-VAE and VAE trained on \texttt{walk} trials. oi-VAE generates physically realistic walking poses while VAE sometimes produces implausible ones.
}
\label{fig:mocap-random-samples}
\end{figure}

One might be concerned that by gaining interpretability, we lose out on expressivity.  However, as we demonstrate in Table~\ref{table:mocap-test-loglik-better} and Figs.~\ref{fig:mocap-reconstruction-samples}-\ref{fig:mocap-random-samples}, the regularization provided by our sparsity-inducing penalty actually leads to as good or better performance across various metrics of model fit.  We first examine oi-VAE and VAE's ability to generalize to held out data.  To examine robustness to different amounts of training data, we consider training on increasing numbers of \texttt{walk} trials and testing on a single heldout example of either \texttt{walk} or \texttt{brisk walk}.  The latter represents an example of data that is a slight variation of what was trained on, whereas the former is a heldout example that is very similar to the training examples.  In Table~\ref{table:mocap-test-loglik-better}, we see the benefit of the regularization in oi-VAE in both test scenarios in the limited data regime.  Not surprisingly, for the full 10 trials, there are little to no differences between the generalization abilities of oi-VAE and VAE (though of course the oi-VAE still provides interpretability).  We highlight that when we have both a limited amount of training data that might not be fully representative of the full possible dataset of interest (e.g., all types of walking), the regularization provided by oi-VAE provides dramatic improvements for generalization.  Finally, in almost all scenarios, the more decomposed oi-VAE $K=16$ setting has better or comparable performance to smaller $K$ settings.

\begin{table}[t]
\caption{Top 3 joints associated with each latent dimension. Grayscale values determined by $\vv{W}_{\cdot,j}^{(g)}$. We see kinematically associated joints associated with each latent dimension.}
\label{table:mocap-topic-model-thing}
\begin{center}
\begin{small}
\begin{tabular}{c|l}
\toprule
\textsc{Dim.} & \textsc{Top 3 joints} \\
\midrule
1 & {\faded{0.1286}left foot},        {\faded{0.1104}left lower leg},   {\faded{0.1096}left upper leg} \\
2 & {\faded{0.2466}head},             {\faded{0.2403}upper neck},       {\faded{0.1849}lower neck} \\
3 & {\faded{0.3261}left thumb},       {\faded{0.3249}left hand},        {\faded{0.0589}left upper arm} \\
4 & {\faded{0.2122}left wrist},       {\faded{0.2002}left upper arm},   {\faded{0.1458}left lower arm} \\
5 & {\faded{0.2052}left lower leg},   {\faded{0.1584}left upper leg},   {\faded{0.1253}right lower leg} \\
6 & {\faded{0.1908}upper back},       {\faded{0.1129}thorax},           {\faded{0.0630}lower back} \\
7 & {\faded{0.0038}left hand},        {\faded{0.0037}left thumb},       {\faded{0.0025}upper back} \\
8 & {\faded{0.1343}head},             {\faded{0.1076}upper neck},       {\faded{0.0925}lower back} \\
9 & {\faded{0.1599}right lower arm},  {\faded{0.1294}right wrist},      {\faded{0.1212}right upper arm} \\
10 & {\faded{0.2303}head},            {\faded{0.2135}upper neck},       {\faded{0.0752}lower neck} \\
11 & {\faded{0.1372}thorax},          {\faded{0.1266}lower back},       {\faded{0.0637}upper back} \\
12 & {\faded{0.2190}left upper leg},  {\faded{0.1832}right foot},       {\faded{0.1798}root} \\
13 & {\faded{0.1558}lower back},      {\faded{0.1393}thorax},           {\faded{0.0601}right upper leg} \\
14 & {\faded{0.2676}thorax},          {\faded{0.2640}upper back},       {\faded{0.2592}lower back} \\
15 & {\faded{0.1248}right upper leg}, {\faded{0.0810}right lower leg},  {\faded{0.0752}left upper leg} \\
16 & {\faded{0.1549}right foot},      {\faded{0.1140}right upper leg},  {\faded{0.0995}left foot} \\
\bottomrule
\end{tabular}
\end{small}
\end{center}
\end{table}

% \begin{table}[t]
% \caption{Top 3 joints associated with each latent dimension. Grayscale values determined by $\vv{W}_{\cdot,j}^{(g)}$. We see kinematically associated joints associated with each latent dimension.}
% \label{table:mocap-topic-model-thing}
% \begin{center}
% \includegraphics[width=0.5\columnwidth]{fig/mocap/joint-table.png}
% \end{center}
% \end{table}

Next, we turn to assessing the learned oi-VAE's generative process relative to that of the VAE.  In Fig.~\ref{fig:mocap-reconstruction-samples} we take our test trial of \texttt{walk}, run each frame through the learned inference network to get a set of latent embeddings $\vv{z}$.  For each such $\vv{z}$, we sample 32 times from $q_\phi(\vv{z}|\vv{x})$ and run each through the generator networks to synthesize a new frame mini-``sequence'', where really the elements of this sequence are the perturbed samples about the embedded test frame.  To fully explore the space of human motion the learned generators can capture, in Fig.~\ref{fig:mocap-random-samples} we sample the latent space at random 100 times from the prior.  For each \emph{unconditional} sample of $\vv{z}$, we pass it through the trained generator to create new frames.  A representative subset of these frames is shown in Fig.~\ref{fig:mocap-random-samples}.  We also show similarly sampled frames from the trained VAE.  A full set of 100 random samples from both VAE and oi-VAE are provided in the Supplement.  Note that, even when trained on the full set of 10 \texttt{walk} trials where we see little to no difference in test log-likelihood between the oi-VAE and VAE, we do see that the learned generator for the oi-VAE is more representative of physically plausible human motion poses.  We attribute this to the fact that the generators of the oi-VAE are able to focus on local correlation structure.

% In summary, these results demonstrate how the oi-VAE provides interpretability while maintaining at least as good expressive power as the VAE in settings where there is a natural notion of groupings of observations.

\begin{table*}[t!]
\caption{\small Test log-likelihood for VAE and oi-VAE trained on 1,2,5, or 10 trials of \texttt{walk} data. Table includes results for a test \texttt{walk} (same as training) or \texttt{brisk walk} trial (unseen in training). Bold numbers indicate the best performance.}
\label{table:mocap-test-loglik-better}
\begin{center}
\begin{small}
\begin{sc}

\begin{tabular}{c}
standard walk
\end{tabular}

\begin{tabular}{l|cccc}
\toprule
\# trials       & 1                 & 2                 & 5             & 10                \\
\midrule
VAE ($K=16$)    & $-3,518$          & $-251$            & $18$          & $\mathbf{114}$    \\
oi-VAE ($K=4$)  & $\mathbf{-2,722}$ & $-214$            & $27$          & $70$              \\
oi-VAE ($K=8$)  & $-3,196$          & $-195$            & $29$          & $75$              \\
oi-VAE ($K=16$) & $-3,550$          & $\mathbf{-188}$   & $\mathbf{31}$ & $108$             \\
\bottomrule
\end{tabular}

\vskip 10pt

\begin{tabular}{c}
brisk walk
\end{tabular}

\begin{tabular}{l|cccc}
\toprule
\# trials       & 1                     & 2                     & 5                         & 10 \\
\midrule
VAE ($K=16$)    & $-723,795$            & $-15,413,445$         & $-19,302,644$             & $-19,303,072$ \\
oi-VAE ($K=4$)  & $-664,608$            & $-13,438,602$         & $\mathbf{-19,289,548}$    & $-19,302,680$ \\
oi-VAE ($K=8$)  & $-283,352$            & $-10,305,693$         & $-19,356,218$             & $-19,302,764$ \\
oi-VAE ($K=16$) & $\mathbf{-198,663}$   & $\mathbf{-6,781,047}$ & $-19,299,964$             & $-19,302,924$ \\
\bottomrule
\end{tabular}

\end{sc}
\end{small}
\end{center}
\end{table*}

\subsection{Magnetoencephalography}
\label{sec:MEG}

Magnetoencephalography (MEG) records the weak magnetic field produced by the brain during cognitive activity with great temporal resolution and good spatial resolution. Analyzing this data holds great promise for understanding the neural underpinnings of cognitive behaviors and for characterizing neurological disorders such as autism. A common step when analyzing MEG data is to project the MEG sensor data into \textit{source-space} where we obtain observations over time on a high-resolution mesh ($\approx$ $5$-$10$K vertices) of the cortical surface~\cite{Gramfort:2013}. The resulting source-space signals likely live on a low-dimensional manifold making methods such as the VAE attractive. However, neuroscientists have meticulously studied particular brain regions of interest and what behaviors they are involved in, so that a key problem is inferring groups of interrelated regions.

\begin{figure}[h]
\centering
\begin{subfigure}{0.45\columnwidth}
  \centering
  \includegraphics[scale=0.45]{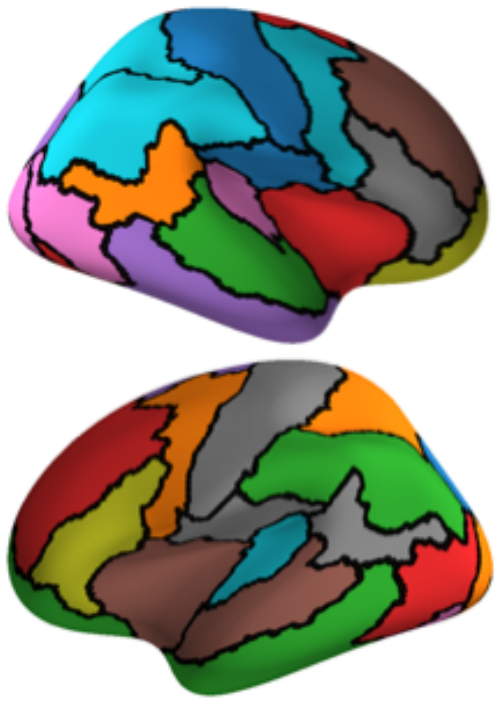}
  \label{fig:group_matrix_brain}
\end{subfigure}
\begin{subfigure}{0.45\columnwidth}
  \centering
  \includegraphics[scale=0.35]{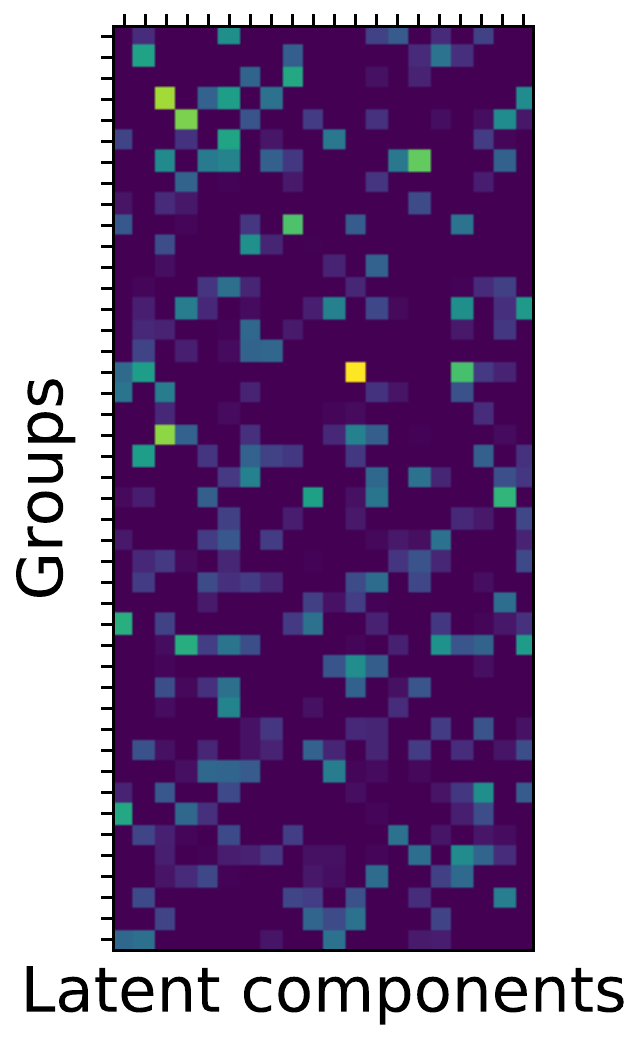}
  \label{fig:group_matrix_z}
\end{subfigure}
\caption{\small (Left) The regions making up the HCP-MMP1 parcellation defining the groups. (Right) Latent-to-group mappings indicate that each latent component influences a sparse set of regions.}
\label{fig:meg_z_matrix}
\end{figure}

We apply our oi-VAE method to infer low-rank represenations of source-space MEG data where the groups are specified as the $\approx 40$ regions defined in the HCP-MMP1 brain parcellation~\cite{Glasser:2016}. See Fig.~\ref{fig:meg_z_matrix}(left). The recordings were collected from a single subject performing an auditory attention task where they were asked to maintain their attention to one of two auditory streams. We use 106 trials each of length 385. We treat each time point of each trial as an i.i.d.\ observation resulting in $\approx 41$K observations. For details on the specification of all considered models, see the Supplement.

For each region we compute the average sensor-space activity over all vertices in the region resulting in 44-dimensional observations. We applied oi-VAE with $K=20$, $\lambda=1$, and Alg.~\ref{alg:vi} for $10,000$ iterations. In Fig.~\ref{fig:meg_z_matrix} we depict the learned group-weights $||\vv{W}^{(g)}_{\cdot,j}||_2$ for all groups $g$ and components $j$.  We observe that each component manifests itself in a sparse subset of the regions. Next, we dig into specific latent components and evaluate whether each influences a subset of regions in a neuroscientifically interpretable manner.

\begin{figure}[t]
\centering
\begin{tabular}{c}
\includegraphics[width=0.95\linewidth]{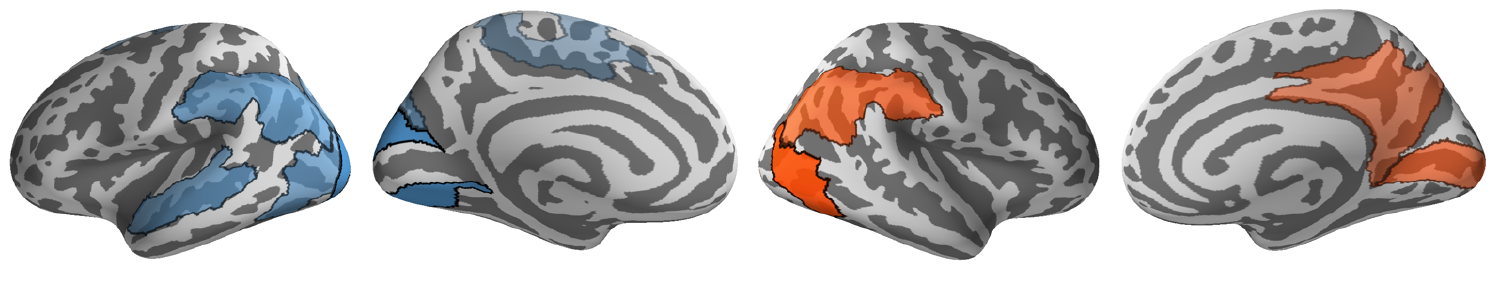} \\
\includegraphics[width=0.95\linewidth]{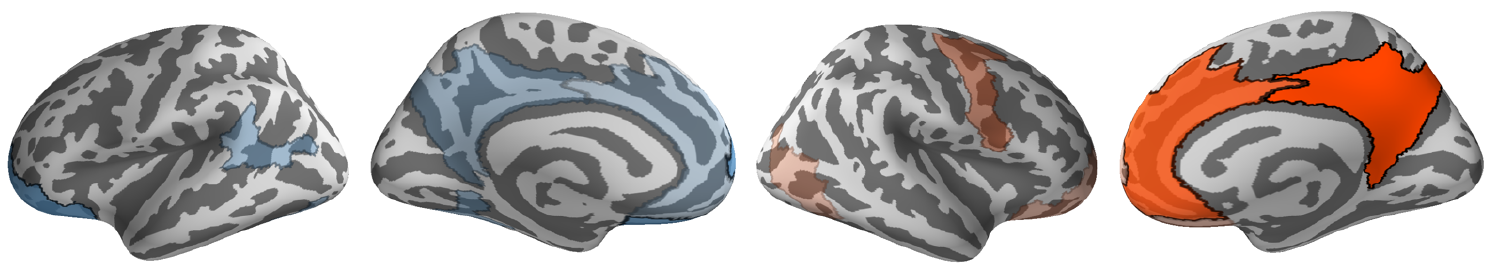}
\end{tabular}
\caption{\small Influence of $\mathbf{z}_6$ (top) and $\mathbf{z}_{15}$ (bottom) on the HCP-MMP1 regions. Active regions (shaded) correspond to the \textit{dorsal attention network} and \textit{default mode network}, respectively.}
\label{fig:meg-components}
\end{figure}

For a given latent component $\vv{z_j}$, the value $||\mathbf{W}^{(g)}_{\cdot,j}||_2$ allows us to interpret how much component $j$ influences region $g$.
We visualize some of these weights for two prominent learned components in Fig.~\ref{fig:meg-components}. Specifically, we find that component 6 captures the regions that make up the \emph{dorsal attention network} pertaining to an auditory spatial task, viz., early visual, auditory sensory areas as well as inferior parietal sulcus and the region covering the right temporoparietal junction~\cite{Lee:2014}. We also find that component 15 corresponds to regions associated with the \emph{default mode network}, viz., medial prefrontal as well as posterior cingulate cortex~\cite{Buckner:2008}. %Since the activations of regions in the default mode network rise and fall together, we may be seeing the push-pull of the attention and default mode networks while the subject engages attends to a stimulus. This hypothesis can be further probed by estimating functional connectivity between the two learned sets of groups.
Again the oi-VAE leads to interpretable results that align with meaningful and previously studied physiological systems.  These systems can be further probed through functional connectivity analysis.  See the Supplement for the analysis of more components.

\section{Conclusion}
We proposed an output interpretable VAE (oi-VAE) that can be viewed as either a nonlinear group latent factor model or as a structured VAE with disentangled latent embeddings.  The approach combines deep generative models with a hierarchical sparsity-inducing prior that leads to our ability to extract meaningful notions of latent-to-observed interactions when the observations are structured into groups.  From this interaction structure, we can infer correlated systems of interaction amongst the observational groups.  In our motion capture and MEG experiments we demonstrated that the resulting systems are physically meaningful.  Importantly, this interpretability does not appear to come at the cost of expressivity, and in our group-structured case can actually lead to improved generalization and generative processes.

In contrast to alternative approaches one might consider for nonlinear group sparse factor analysis, leveraging the amortized inference associated with VAEs leads to computational efficiencies.  We see even more significant gains through our proposed collapsed objective.  The proximal updates we can apply lead to real learned sparsity.

We note that nothing fundamentally prevents applying this architecture to other generative models \textit{du jour}. Extending this work to GANs, for example, should be straightforward.  Furthermore, one could consider combining this framework with sparsity inducing priors on $\vv{z}$ to discourage redundant latent dimensions. Oy-vey!

\section*{Acknowledgements}
This work was supported by ONR Grant N00014-15-1-2380, NSF CAREER Award IIS-1350133, NSF CRCNS Grant NSF-IIS-1607468, and AFOSR Grant FA9550-1-1-0038. The authors also gratefully acknowledge the support of NVIDIA Corporation for the donated GPU used for this research.

\pagebreak

\bibliographystyle{icml2018}
\bibliography{main}

\pagebreak

\appendix
% \section{Supplementary material}
% %% Language and font encodings
% \usepackage[english]{babel}
% \usepackage[utf8x]{inputenc}
% \usepackage[T1]{fontenc}

% \usepackage{booktabs} % for professional tables

% %% Sets page size and margins
% \usepackage[top=3cm,bottom=3cm,left=3cm,right=3cm,marginparwidth=1.75cm]{geometry}

% %% Useful packages
% \usepackage{amsmath}
% \usepackage{amsfonts}
% \usepackage{graphicx}
% \usepackage[colorinlistoftodos]{todonotes}
% \usepackage[colorlinks=true, allcolors=blue]{hyperref}
% \usepackage{subcaption}

\section{Synthetic bars data}
\begin{figure*}[h]
\centering
\begin{subfigure}{0.45\columnwidth}
  \centering
  \includegraphics[width=\linewidth]{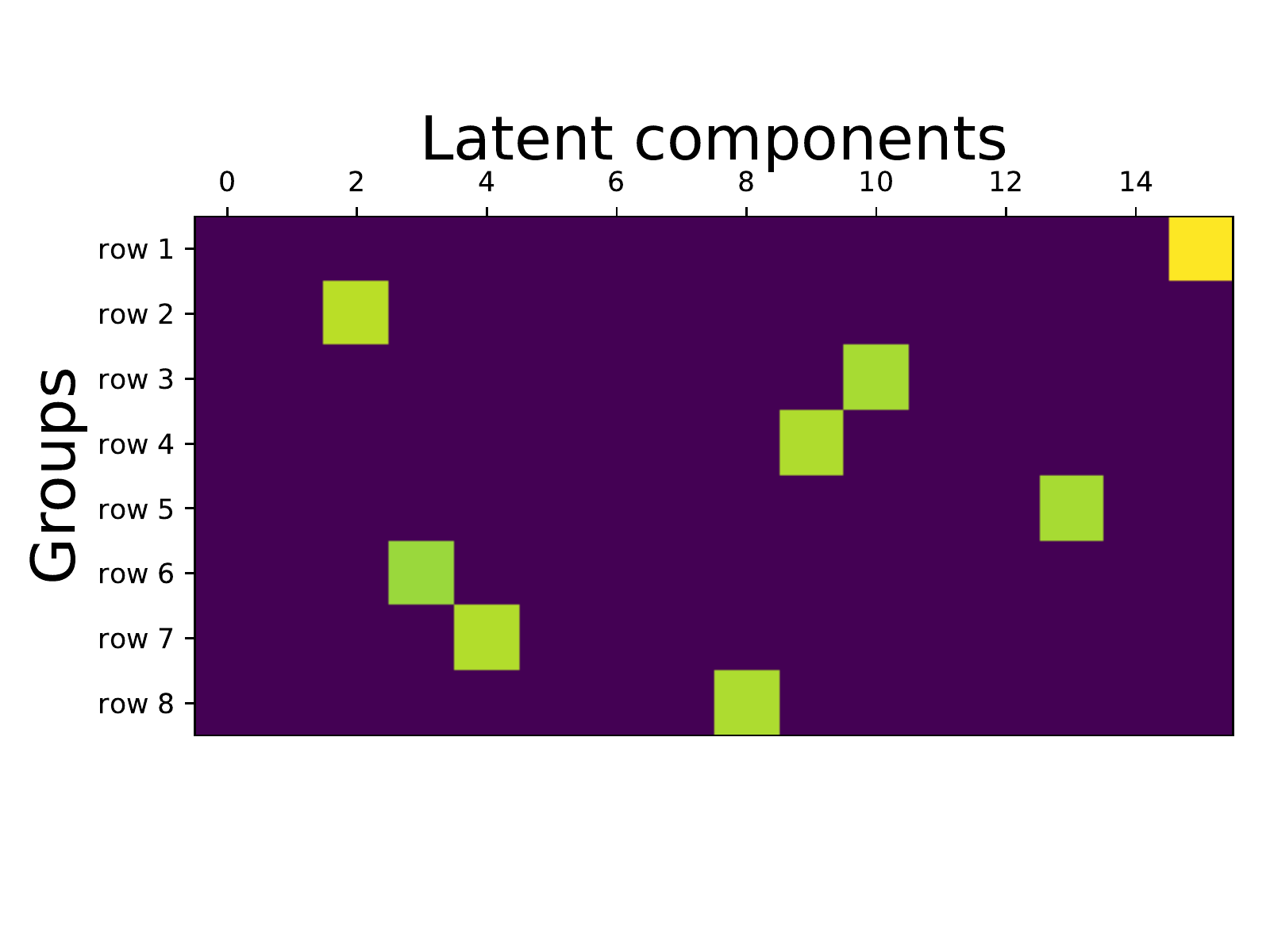}
  \caption{oi-VAE with $\lambda=1$}
\end{subfigure}
\begin{subfigure}{0.45\columnwidth}
  \centering
  \includegraphics[width=\linewidth]{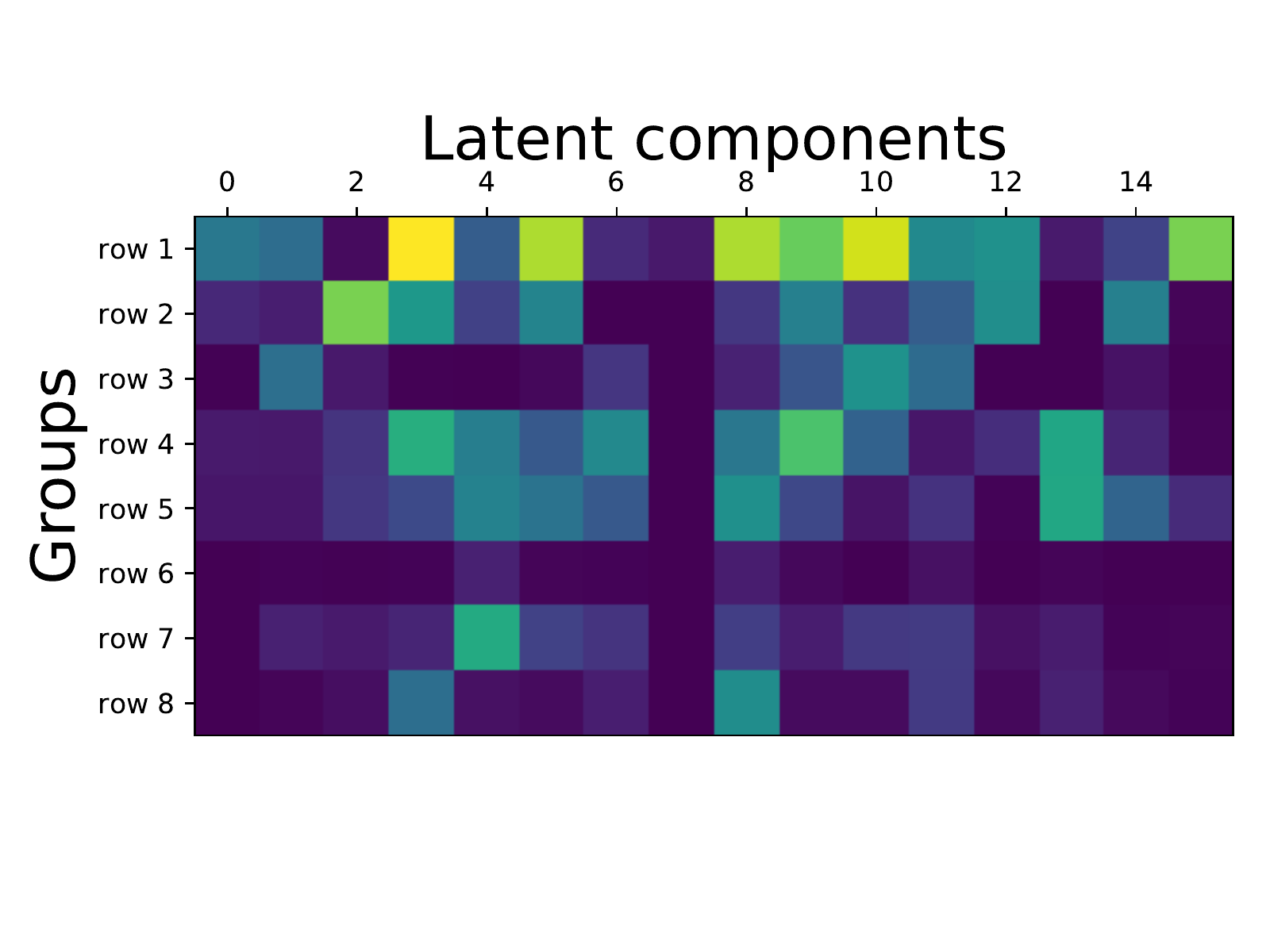}
  \caption{No prior on the $\mathcal{W}$ or $\theta$.}
\end{subfigure}
\caption{Results with latent dimension $K=16$ when the effective dimensionality of the data is only $8$. Clearly the oi-VAE has learned to use only the sparse set of $z_i$'s that are necessary to explain the data.}
\label{fig:bars-data-big-K}
\end{figure*}

In addition to the evaluations shown in the paper, we evaluated oi-VAE when the number of latent dimensions $K$ is greater than necessary to fully explain the data. In particular we sample the same $8 \times 8$ images but use $K=16$. See Figure \ref{fig:bars-data-big-K}. Train and test log likelihoods for the model are given in Table \ref{table:bars-data-big-K}.

\paragraph{Experimental details}
For all our synthetic data experiments we sampled 2,048 $8 \times 8$ images with exactly one bar present uniformly at random. The activated bar was given a value of $0.5$, inactive pixels were given values of zero. White noise was added to the entire image with standard deviation 0.05. We set $p = 1$ and $\lambda = 1$.
\begin{itemize}
\item Inference model:
\begin{itemize}
\item $\mu(\mathbf{x}) = W_1 \mathbf{x} + b_1$.
\item $\sigma(\mathbf{x}) = \exp(W_2 \mathbf{x} + b_2$).
\end{itemize}
\item Generative model:
\begin{itemize}
\item $\mu(\mathbf{z}) = W_3 \mathbf{z} + b_3$.
\item $\sigma = \exp(b_4)$.
\end{itemize}
\end{itemize}
We ran Adam on the inference and generative net parameters with learning rate $1e-2$. Proximal gradient descent was run on $\mathcal{W}$ with learning rate $1e-4$. We used a batch size of 64 sampled uniformly at random at each iteration and ran for 20,000 iterations.

\begin{table}[h]
\caption{Train and test log likelihoods on the synthetic bars data when $K$ is larger than necessary.}
\label{table:bars-data-big-K}
% \vskip 0.15in
\begin{center}
\begin{small}
\begin{tabular}{l|cc}
\toprule
\textsc{Model} & \textsc{Train log likelihood} & \textsc{Test log likelihood} \\
\midrule
$\lambda=1$ & $\mathbf{99.9325}$ & $\mathbf{100.1394}$ \\
$\lambda=0$ and no $\theta$ prior & $95.0687$ & $95.4285$ \\
\bottomrule
\end{tabular}
\end{small}
\end{center}
% \vskip -0.1in
\end{table}

% $\mathbf{W}^{(g)}_{\boldsymbol{\cdot}, j}$

\section{Motion capture results}
\begin{figure*}[h]
\centering
\begin{subfigure}{0.85\columnwidth}
  \centering
  \includegraphics[width=\linewidth]{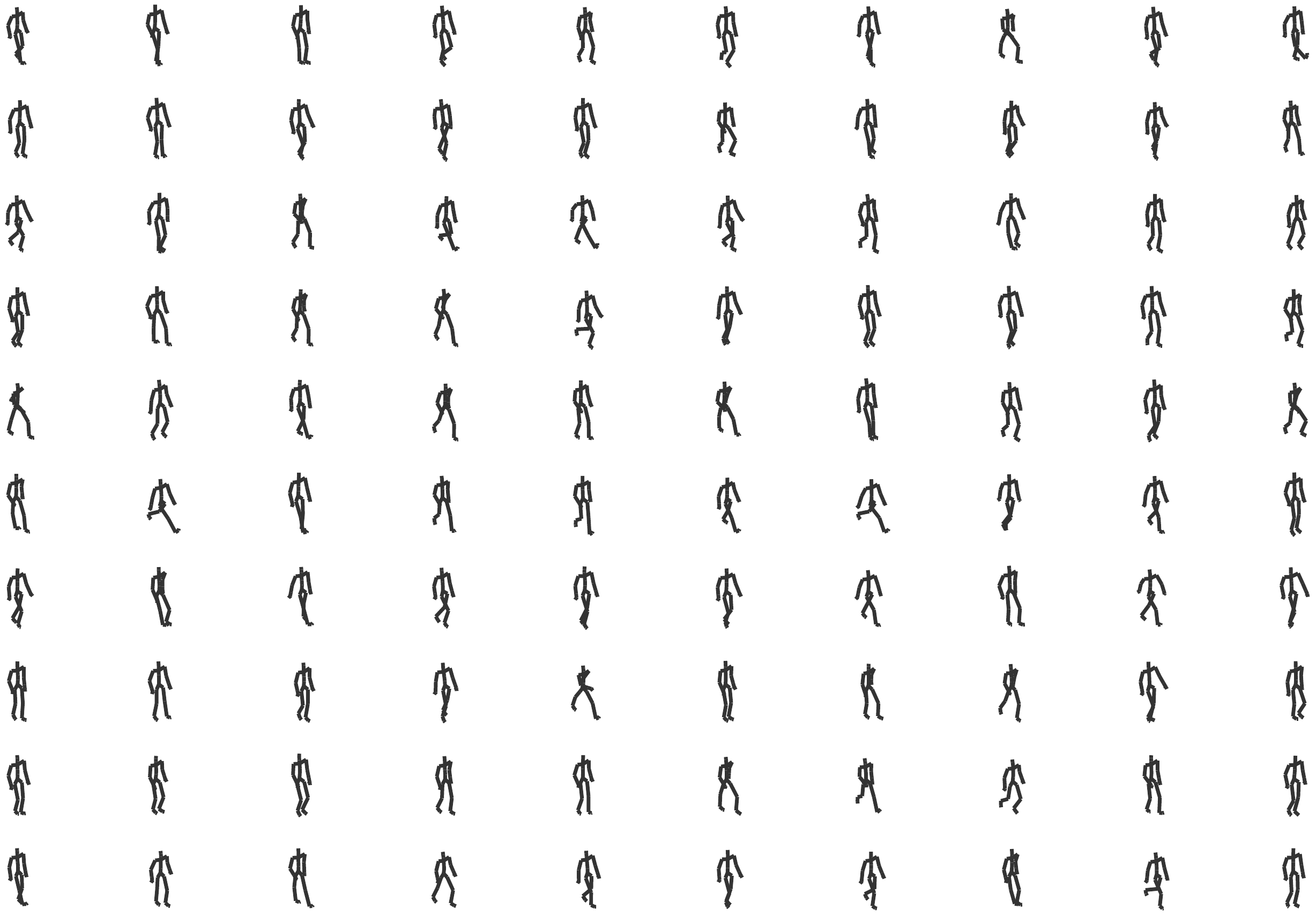}
  \caption{}
\end{subfigure}
\begin{subfigure}{0.85\columnwidth}
  \centering
  \includegraphics[width=\linewidth]{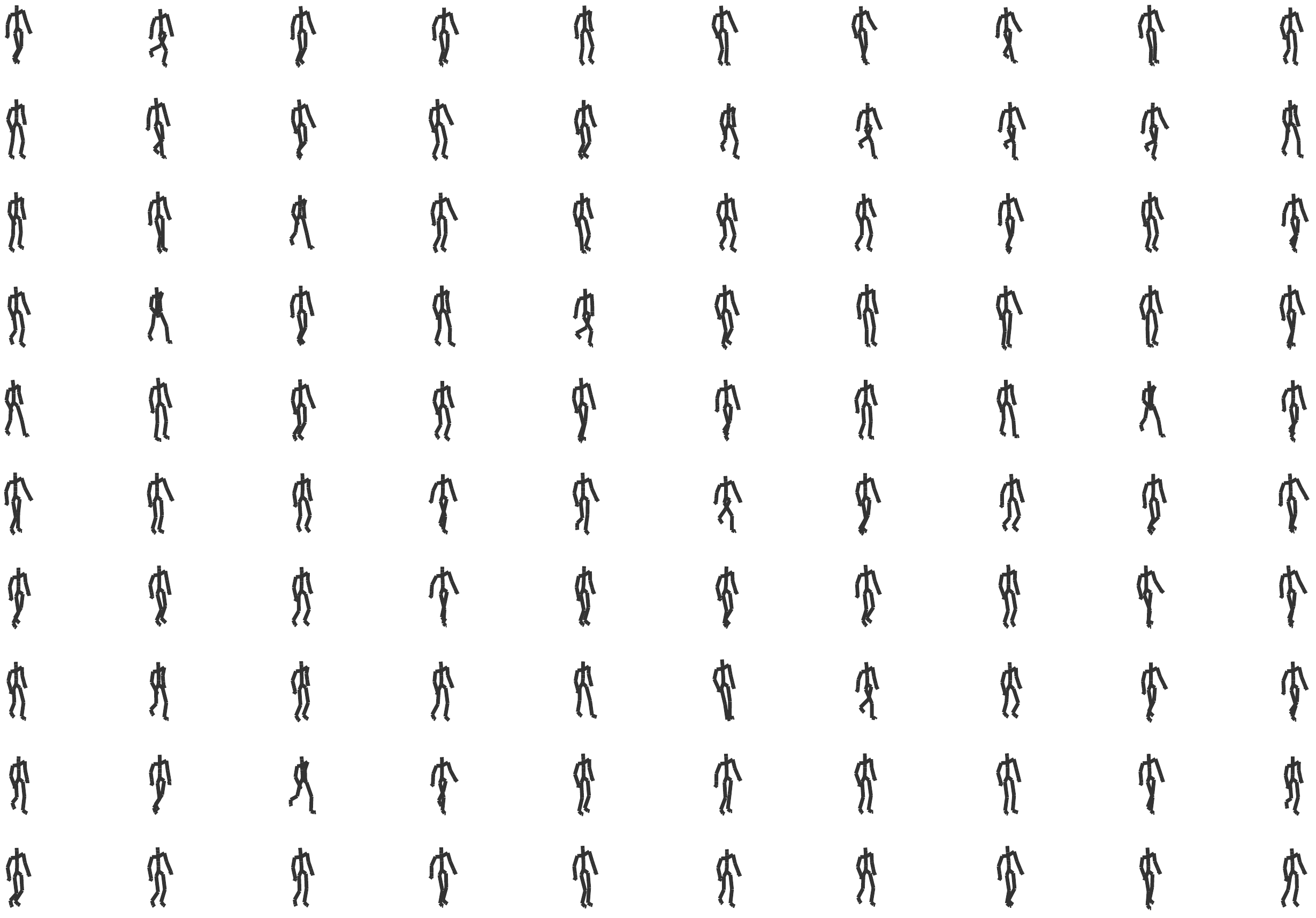}
  \caption{}
\end{subfigure}
\caption{Samples from the (a) VAE and (b) oi-VAE models. The VAE produces a number of poses apparently inspired by the \href{https://www.youtube.com/watch?v=iV2ViNJFZC8}{Ministry of Silly Walks}. Some others are even physically impossible. In contrast, results from the oi-VAE are all physically plausible and appear to be representative of walking. Full scale images will be made available on the author's website.}
\label{fig:vae-and-oivae-100-samples}
\end{figure*}

Multiple samples from both the VAE and oi-VAE are shown in Figure \ref{fig:vae-and-oivae-100-samples}.

\paragraph{Experimental details} We used data from subject 7 in the CMU Motion Capture Database. Trials 1-10 were used for training. Trials 11 and 12 were left out to form a test set. Trial 11 is a standard \texttt{walk} trial. Trial 12 is a \texttt{brisk walk} trial. We set $p=8$ and $\lambda=1$.
\begin{itemize}
\item Inference model:
\begin{itemize}
\item $\mu(\mathbf{x}) = W_1 \mathbf{x} + b_1$.
\item $\sigma(\mathbf{x}) = \exp(W_2 \mathbf{x} + b_2$).
\end{itemize}
\item Generative model:
\begin{itemize}
\item $\mu(\mathbf{z}) = W_3 \tanh(\mathbf{z}) + b_3$.
\item $\sigma = \exp(b_4)$.
\end{itemize}
\end{itemize}
We ran Adam on the inference and generative net parameters with learning rate $1e-3$. Proximal gradient descent was run on $\mathcal{W}$ with learning rate $1e-4$. We used a batch size of 64 with batches shuffled before every epoch. Optimization was run for 1,000 epochs.

\section{MEG Analysis}
We present the three most prominent components determined by summing $||\mathbf{W}^{(g)}_{\cdot,j}||_2$ over all groups $g$.
These components turn out to be harder to interpret than some of the others presented indicating that the norm of the group-weights may not be the best notion to determine interpretable components.
However, this perhaps is not surprising with neuroimaging data. In fact, the strongest components inferred when applying PCA or ICA to neuroimaging data usually correspond to physiological artifacts such as eye movement or cardiac activity~\cite{Uusitalo:1997}.

We depict the three most prominent latent components according to the group weights. We also depict component 7 which corresponds to the spatial attentional network that consists of a mix of auditory and visual regions. This arises because the auditory attentional network taps into the visual network.

\paragraph{Experimental details} We set $p=10$ and $\lambda=10$. The inference net was augmented with a hidden layer of 256 units.
\begin{itemize}
\item Inference model:
\begin{itemize}
\item $\mu(\mathbf{x}) = W_2 \mathrm{relu}(W_1 \mathbf{x} + b_1) + b_2$.
\item $\sigma(\mathbf{x}) = \exp(W_3 \mathrm{relu}(W_1 \mathbf{x} + b_1) + b_3)$.
\end{itemize}
\item Generative model:
\begin{itemize}
\item $\mu(\mathbf{z}) = W_3 \tanh(\mathbf{z}) + b_3$.
\item $\sigma = \exp(b_4)$.
\end{itemize}
\end{itemize}
We ran Adam on the inference and generative net parameters with learning rate $1e-3$. Proximal gradient descent was run on $\mathcal{W}$ with learning rate $1e-6$. We used a batch size of 256 with batches shuffled before every epoch. Optimization was run for 40 epochs.

\begin{figure*}[h!]
% \vskip 0.2in
\centering
\begin{subfigure}{\columnwidth}
  \centering
  \includegraphics[width=.4\linewidth]{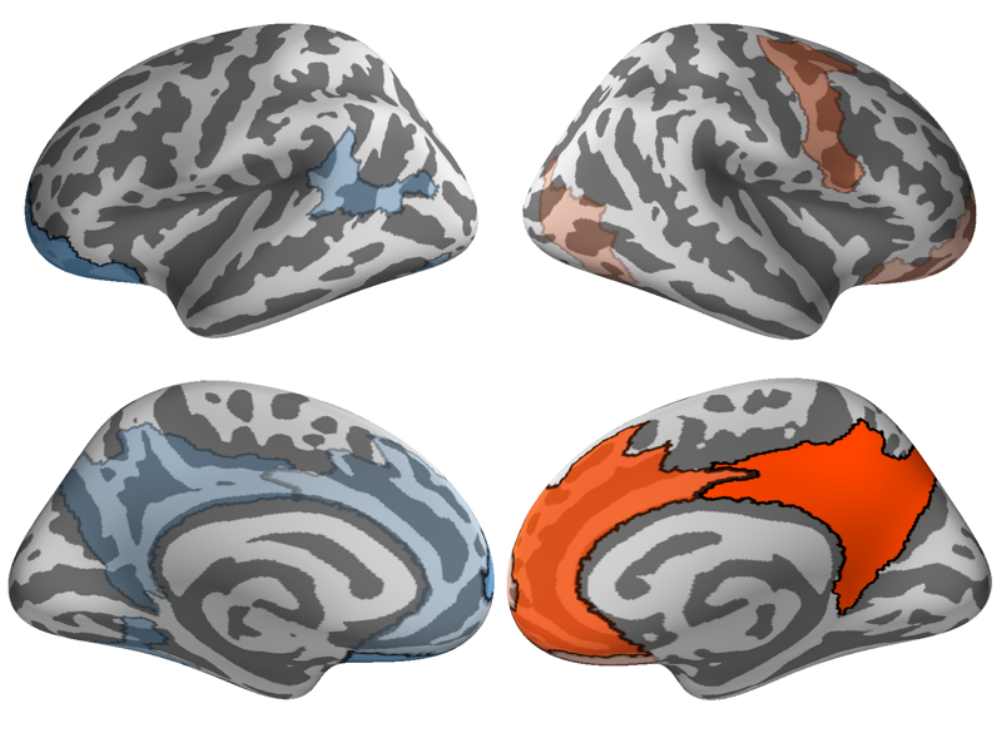}
  \caption{Component 15: Default Mode Network}
  \label{fig:z_15_dmn}
\end{subfigure}
\caption{\small The projections of two components of $\mathbf{z}$ onto the regions of the HCP-MMP1 parcellation. The regions with the ten largest weights are shaded (blue in the left hemisphere, red in the right hemisphere) with opacity indicating the strength of the weight. Component 15 corresponds to the default mode network.}
% \label{fig:meg-components}
\end{figure*}

\begin{figure*}[h]
% \vskip 0.2in
\centering
\begin{subfigure}{0.45\columnwidth}
  \centering
  \includegraphics[width=\linewidth]{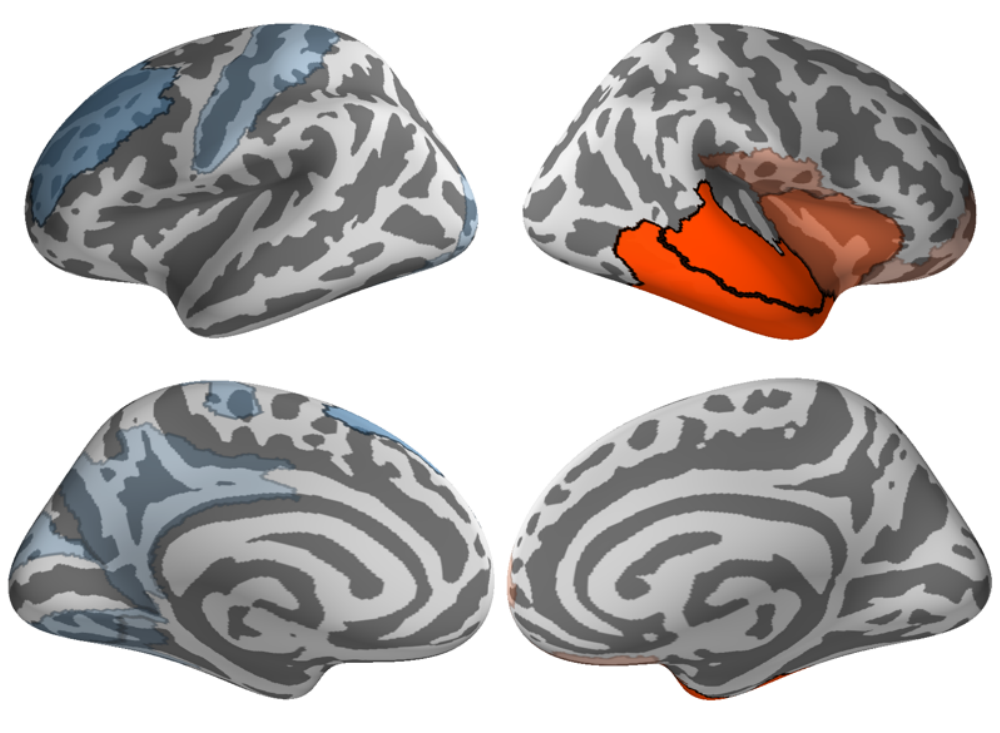}
  \caption{Component 2.}
%   \label{fig:z_2_dan}
\end{subfigure}
\begin{subfigure}{0.45\columnwidth}
  \centering
  \includegraphics[width=\linewidth]{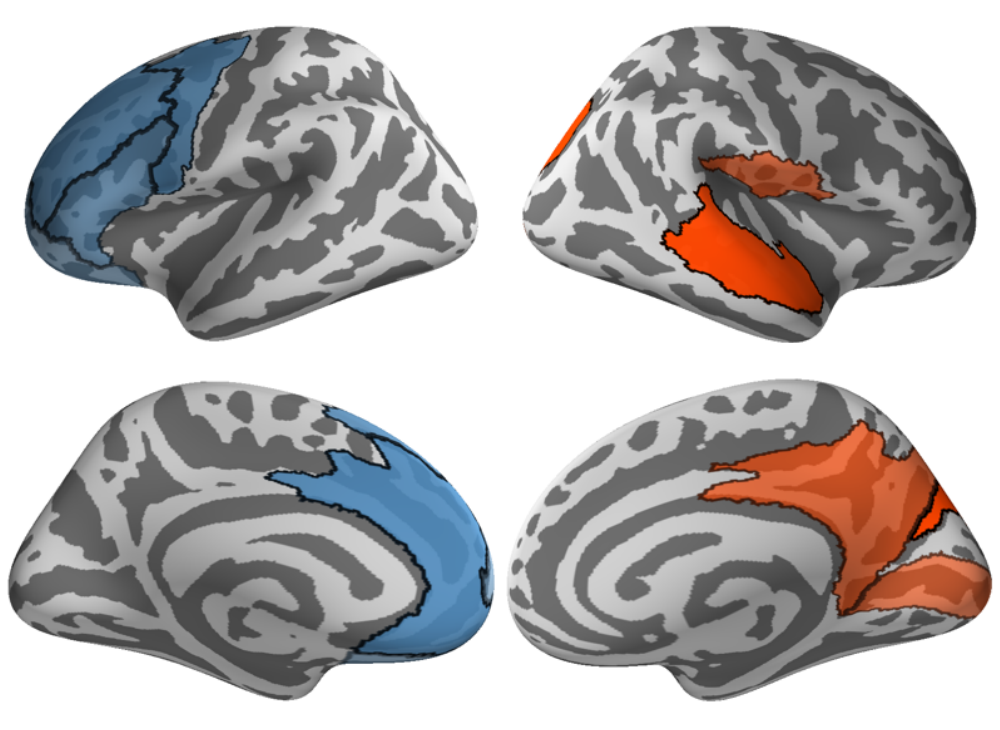}
  \caption{Component 5.}
  \label{fig:z_5_dmn}
\end{subfigure}
\caption{Component 2 has the largest aggregate group weight and component 5 has the second largest.}
% \label{fig:meg-components}
\end{figure*}

\begin{figure*}[h]
% \vskip 0.2in
\centering
\begin{subfigure}{0.45\columnwidth}
  \centering
  \includegraphics[width=\linewidth]{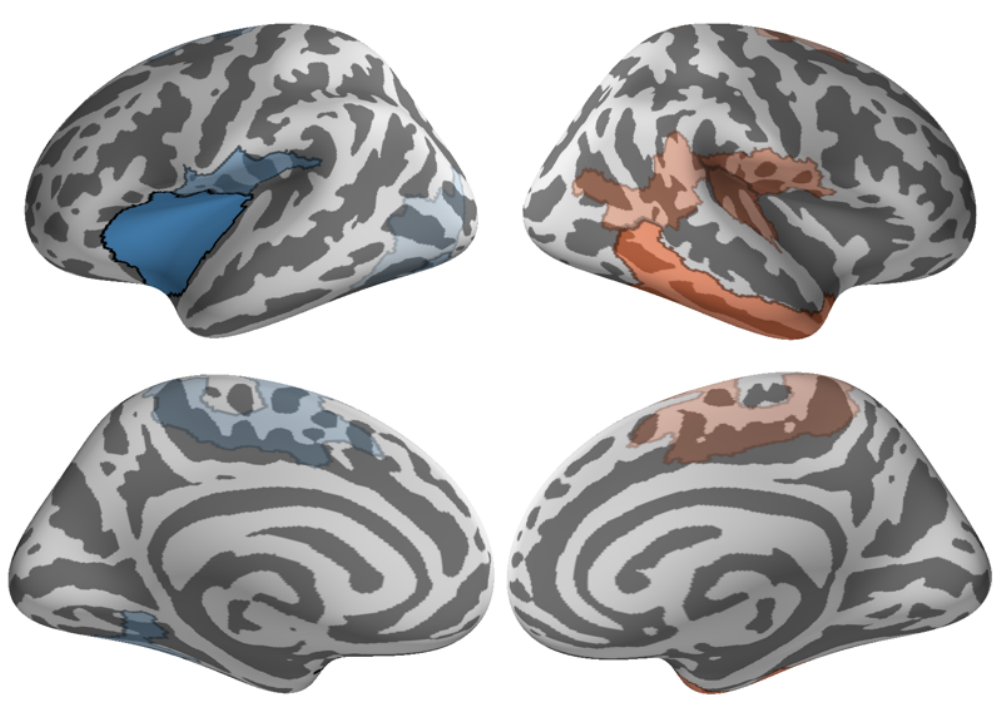}
  \caption{Component 11}
  \label{fig:z_11_dan}
\end{subfigure}
\begin{subfigure}{0.45\columnwidth}
  \centering
  \includegraphics[width=\linewidth]{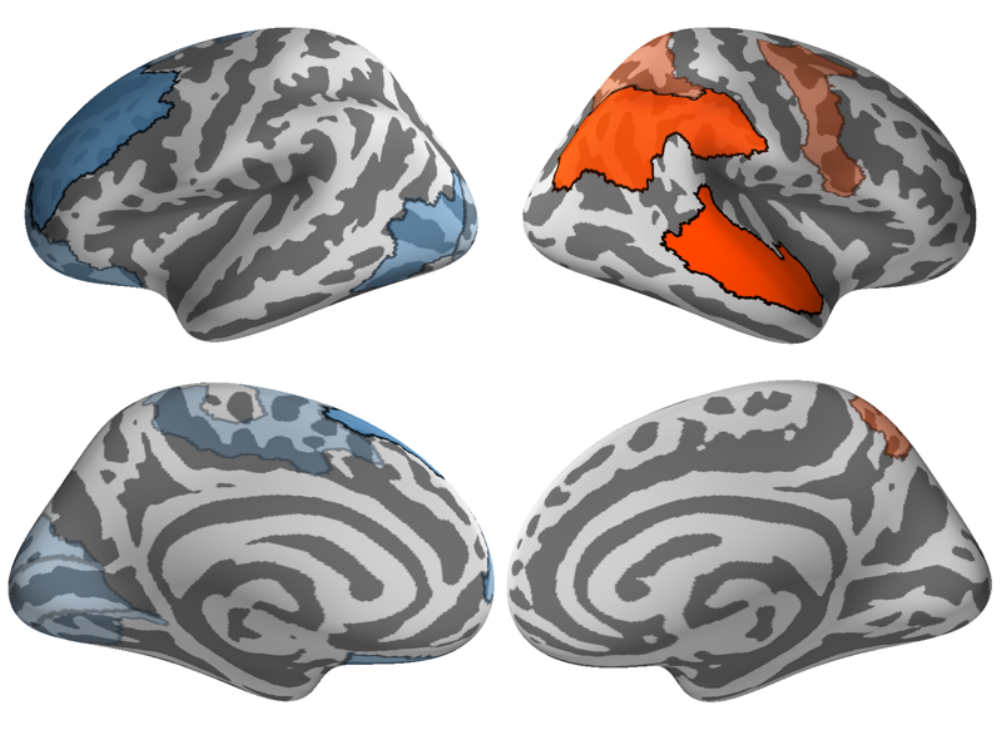}
  \caption{Component 7}
%   \label{fig:z_7_dmn}
\end{subfigure}
\caption{Component 11 resembles the ventral stream and has the third largest aggregate group weight. Component 7 has a smaller aggregate group weight but corresponds to the spatial attentional network.}
% \label{fig:meg-components}
\end{figure*}

\section{Common experimental details}
We found that it was crucial to throttle the variance of the posterior approximation in order stabilize training in the initial stages of optimization for both the VAE and oi-VAE. We did so by multiplying the outputted standard deviations by $0.1$ for the first 25 epochs and then resumed training normally after that point. A small $1e-3$ factor was added to all of the outputted standard deviations in order to promote numerical stability when calculating gradients.

In all of our experiments we estimated $\mathbb{E}_{q_\phi(\mathbf{z} | \mathbf{x})} \left[ \log p(x | \mathbf{z}, \mathcal{W}, \theta) \right]$ with one sample. We experimented with using more samples but did not observe any significant benefit from doing so.

\end{document}